\documentstyle[psfig]{llncs}
\begin{document}
\begin{center}
{\large {\bf A Numerical Example on the\\ Principles of Stochastic
Discrimination}} 
\vspace{0.1in}\\
Tin Kam Ho \vspace{0.025in}\\
Bell Laboratories, Lucent Technologies \vspace{0.05in}\\
July 15, 2003 \vspace{0.5in}\\
{\bf Abstract}
\vspace{0.1in}\\
\parbox{4.5in}{
Studies on ensemble methods for classification suffer from the
difficulty of modeling the complementary strengths of the 
components.  Kleinberg's theory of stochastic discrimination (SD)
addresses this rigorously via mathematical notions of enrichment, 
uniformity, and projectability of an ensemble.  We explain these concepts
via a very simple numerical example that captures the 
basic principles of the SD theory and method.  We focus on a fundamental
symmetry in point set covering that is the key observation
leading to the foundation of the theory.  We believe a better
understanding of the SD method will lead to developments of
better tools for analyzing other ensemble methods.
}
\end{center}

\section{Introduction}

Methods for classifier combination, or ensemble learning, can be divided
into two categories: 1) {\em decision optimization} methods that try
to obtain {\em consensus} among a {\em given} set of classifiers to
make the best decision; 2) {\em coverage optimization} methods that
try to {\em create} a set of classifiers that work best with a {\em
fixed} decision combination function to cover all possible cases.

Decision optimization methods rely on the assumption that the given
set of classifiers, typically of a small size, contain sufficient
expert knowledge about the application domain, and each of them 
excels in a subset of all possible input.  A decision combination
function is chosen or trained to exploit the individual strengths
while avoiding their weaknesses.  Popular combination functions
include majority/plurality votes\cite{kn:lam97}, sum/product
rules\cite{kn:kittler98}, rank/confidence
score combination\cite{kn:ho94}, and probabilistic methods\cite{kn:huang95}.
While numerous successful applications of these methods have been
reported,  the joint capability of the classifiers sets an intrinsic
limitation on decision optimization that the combination functions
cannot overcome.  A challenge in this approach is to find out the
``blind spots'' of the ensemble and to obtain a classifier that covers
them.

Coverage optimization methods use an automatic and systematic
mechanism to generate new classifiers with the hope of covering all
possible cases. A fixed, typically simple, function is used for
decision combination.  This can take the form of training set
subsampling, such as stacking \cite{kn:wolpert92},
bagging\cite{kn:breiman96}, and boosting\cite{kn:freund96}, feature
subspace projection\cite{kn:ho98}, superclass/subclass
decomposition\cite{kn:dietterich95b}, or other forms of random perturbation of
the classifier training procedures \cite{kn:hansen90}.
Open questions in these methods are
1) how many classifiers are enough?  2) what kind of differences
among the component classifiers yields the best combined accuracy?
3) how much limitation is set by the form of the component
classifiers?

Apparently both categories of ensemble methods run into some dilemma.
Should the component classifiers be weakened in order to achieve a
stronger whole?  Should some accuracy be sacrificed for the known
samples to obtain better generalization for the unseen cases?  Do we
seek agreement, or differences among the component classifiers?

A central difficulty in studying the performance of these ensembles
is how to model the complementary strengths among the classifiers.  Many
proofs rely on an assumption of statistical independence of
component classifiers' decisions.  But rarely is there any attempt to
match this assumption with observations of the decisions.  Often,
global estimates of the component classifiers' accuracies are used in
their selection, while in an ensemble what matter more are the local
estimates, plus the relationship between the local accuracy
estimates on samples that are close neighbors in the feature
space.\footnote{there is more discussion on these difficulties in a
recent review\cite{kn:ho02}.} 

Deeper investigation of these issues leads back to three major concerns
in choosing classifiers: discriminative power, use of complementary
information, and generalization power.  A complete theory on ensembles
must address these three issues simultaneously.  Many current
theories rely,  either explicitly or implicitly, on ideal assumptions
on one or two of these issues, or have them omitted entirely,  and are
therefore incomplete.

Kleinberg's theory and method of stochastic discrimination
(SD)\cite{kn:kleinberg90} \cite{kn:kleinberg96} is the first attempt
to explicitly address these issues simultaneously from a mathematical
point of view.   In this theory, rigorous notions are made for
discriminative power, complementary information, and generalization
power of an ensemble.  A fundamental symmetry is observed between
the probability of a fixed model covering a point in a given set and
the probability of a fixed point being covered by a model in a given
ensemble.  The theory establishes that,  these three conditions
are sufficient for an ensemble to converge, with increases in its size, to
the most accurate classifier for the application.

Kleinberg's analysis uses a set-theoretic abstraction to remove all the
algorithmic details of classifiers, features, and training procedures.
It considers only the classifiers' decision regions in the form of point
sets, called weak models, in the feature space.  A collection of
classifiers is thus just a sample from the power set of the feature
space.   If the sample satisfies a uniformity condition, i.e., if its
coverage is unbiased for any local region of the feature space, then a
symmetry is observed between two probabilities (w.r.t. the feature
space and w.r.t. the power set, respectively)  of the same event that a
point of a particular class is covered by a component of the sample.
Discrimination between classes is achieved by requiring some minimum
difference in each component's inclusion of points of different
classes,  which is trivial to satisfy.  By way of this symmetry,
it is shown that if the sample of weak models is large,  the
discriminant function, defined on the coverage of the models on a
single point and the class-specific differences within each model,
converges to poles distinct by class with diminishing variance.

We believe that this symmetry is the key to the discussions on classifier
combination.  However,  since the theory was developed from a fresh,
original, and independent perspective on the problem of learning,  there
have not been many direct links made to the existing theories.
As the concepts are new, the claims are high, the published
algorithms appear simple,  and the details of more sophisticated
implementations are not known, the method has been poorly understood
and is sometimes referred to as mysterious. 

It is the goal of this lecture to illustrate the basic concepts
in this theory and remove the apparent mystery.
We present the principles of stochastic discrimination
with a very simple numerical example.  The example is so chosen
that all computations can be easily traced step-by-step by hand or with
very simple programs.  We use Kleinberg's notation wherever
possible to make it easier for the interested readers to follow up on
the full theory in the original papers.
The emphasis in this note is on explaining the concepts of
uniformity and enrichment, and the behavior of the discriminant when
both conditions are fulfilled.  For the details of the mathematical
theory and outlines of practical algorithms, please refer to the original
publications\cite{kn:kleinberg90}\cite{kn:kleinberg96}\cite{kn:kleinberg00}
\cite{kn:kleinberg00b}.

\section{Symmetry of Probabilities Induced by Uniform Space Covering}

The SD method is based on a fundamental symmetry in point set covering.
To illustrate this symmetry,  we begin with a simple observation.
Consider a set $S=\{a, b, c\}$ 
and all the subsets with two elements $s_1=\{a,b\}$, $s_2=\{a,c\}$,
and $s_3=\{b,c\}$.  By our choice,  each of these subsets has
captured $2/3$ of the elements of $S$.  We call this ratio $r$.  Let
us now look at each member of $S$, and check how many of these three
subsets have included that member.  For example, $a$ is in two of
them, so we say $a$ is captured by $2/3$ of these subsets.  We will
obtain the same value $2/3$ for all elements of $S$.  This value is
the same as $r$. This is a consequence of the fact that we have used
all such 2-member subsets and we have not biased this collection
towards any element of $S$.  With this observation, we begin a larger example.

Consider a set of 10 points in a one-dimensional feature space $F$.
Let this set be called $A$.  Assume that $F$ contains only points in
$A$ and nothing else.  Let each point in $A$ be identified as $q_0$,
$q_1$, ..., $q_9$ as follows. 
\begin{center}
\begin{tabular}{cccccccccc}
.&.&.&.&.&.&.&.&.&.\\
$q_0$ &$q_1$ &$q_2$ &$q_3$ &$q_4$ &$q_5$ &$q_6$ &$q_7$ &$q_8$ &$q_9$\\
\end{tabular}
\end{center}

Now consider the subsets of $F$.  Let the collection of
all such subsets be ${\cal M}$, which is the power set of $F$.
We call each member $m$ of ${\cal M}$ a {\em model},  and we restrict our
consideration to only those models that contain 5 points in $A$,
therefore each model has a size that is 0.5 of the size of $A$.  Let
this set of models be called $M_{0.5,A}$.  Some members of $M_{0.5,A}$
are as follows. 
\begin{center}
\begin{tabular}{ccccc}
\{$q_0$, &$q_1$, &$q_2$, &$q_3$, &$q_4$ \}\\
\{$q_0$, &$q_1$, &$q_2$, &$q_3$, &$q_5$ \}\\
\{$q_0$, &$q_1$, &$q_2$, &$q_3$, &$q_6$ \}\\
...&&&&\\
\end{tabular}
\end{center}

There are $C(10,5) = 252$ members in $M_{0.5,A}$.  Let $M$ be a
pseudo-random permutation of members in $M_{0.5,A}$ as listed
in Table \addtocounter{table}{+1}\thetable\addtocounter{table}{-1}
in the Appendix.  We identify models in this sequence by a single
subscript such that $M = m_1, m_2, ..., m_{252}$.
We expand a collection $M_t$ by including more and more members of
$M_{0.5,A}$ in the order of the sequence $M$ as follows.
$M_1=\{m_1\}$, $M_2=\{m_1,m_2\}$, ..., $M_t=\{m_1,m_2,...m_t\}$.

Since each model covers some points in $A$,  for each member $q$ in $A$,
we can count the number of models in $M_t$ that include $q$,
call this count $N(q,M_t)$, and calculate the ratio of this count over
the size of $M_t$, call it $Y(q,M_t)$.  That is, $Y(q,M_t) =
Prob_{\cal M} (q\in m | m \in M_t)$.  As $M_t$ expands, this ratio
changes and we show these changes for each $q$ in Table
\addtocounter{table}{+2}\thetable\addtocounter{table}{-2} in the Appendix.
The values of $Y(q,M_t)$ are plotted in 
Figure \addtocounter{figure}{1}\thefigure\addtocounter{figure}{-1}.
As is clearly visible in the Figure, the values of $Y(q,M_t)$ converge
to 0.5 for each $q$.  Also notice that because of the randomization,
we have expanded $M_t$ in a way that $M_t$ is not biased towards any
particular $q$,  therefore the values of $Y(q,M_t)$ are similar
after $M_t$ has acquired a certain size (say, when $t=80$).
When $M_t$=$M_{0.5,A}$, every point $q$ is covered by the
same number of models in $M_t$, and their values of $Y(q,M_t)$ are
identical and is equal to 0.5,  which is the ratio of the size of each
$m$ relative to $A$ (recall that we always include 5 points from $A$
in each $m$).

Formally, when $t=252$, $M_t = M_{0.5,A}$, from the
perspective of a fixed $q$, the probability of it being contained in a
model $m$ from $M_t$ is 
\[ Prob_{\cal M} (q \in m | m \in M_{0.5,A}) = 0.5. \]
We emphasize that this probability is a measure in the space ${\cal M}$
by writing the probability as $Prob_{\cal M}$.
On the other hand,  by the way each $m$ is constructed, we know that
from the perspective of a fixed $m$,
\[ Prob_F (q \in m| q \in A) ~=~ 0.5. \]
Note that this probability is a measure in the space $F$.
We have shown that these two probabilities, w.r.t. two different
spaces, have identical values.  In other words,
let the membership function of $m$ be $C_m(q)$, i.e., $C_m(q) =
1 ~{\rm iff}~ q \in m$,  the random variables
$\lambda q C_m(q)$ and $\lambda m C_m(q)$ have the same probability
distribution, when $q$ is restricted to $A$ and $m$ is restricted to
$M_{0.5,A}$.  This is because both variables can have values that are either
1 or 0, and they have the value 1 with the same probability (0.5 in this case).
This symmetry arises from the fact that the collection of models $M_{0.5,A}$
covers the set $A$ uniformly, i.e.,  since we have used all members of
$M_{0.5,A}$, each point $q$ have the same chance to be included
in one of these models.   If any two points in a set $S$ have
the same chance to be included in a collection of models,
we say that this collection is $S$-uniform.  It can be shown, by a simple
counting argument, that uniformity leads to the symmetry of 
$Prob_{\cal M} (q \in m | m \in M_{0.5,A})$ and
$Prob_F (q \in m| q \in A)$, and hence
distributions of $\lambda q C_m(q)$ and $\lambda m C_m(q)$.

The observation and utilization of
this duality are central to the theory of stochastic discrimination.
A critical point of the SD method is to enforce such a uniform
cover on a set of points.  That is, to construct a collection of
models in a balanced way so that the uniformity (hence the duality) is
achieved without exhausting all possible models from the space.

\begin{figure}[h]
\centering
\begin{tabular}{c}
\psfig{file=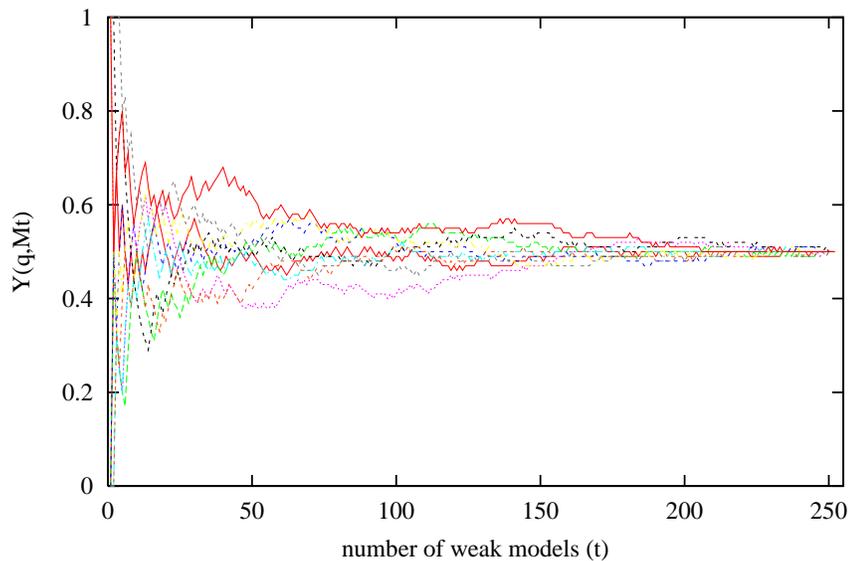,height=3in,clip=}\\
\end{tabular}
\caption{Plot of $Y(q,M_t)$ versus $t$.  Each line represents the
trace of $Y(q,M_t)$ for a particular $q$ as $M_t$ expands.}
\end{figure}

\section{Two-Class Discrimination}

Let us now label each point $q$ in $A$ by one of two classes $c_1$ 
(marked by ``x'') and $c_2$ (marked by ``o'') as follows.

\begin{center}
\begin{tabular}{cccccccccc}
x&x&x&o&o&o&o&x&x&o\\
$q_0$ &$q_1$ &$q_2$ &$q_3$ &$q_4$ &$q_5$ &$q_6$ &$q_7$ &$q_8$ &$q_9$\\
\end{tabular}
\end{center}

This gives a training set $TR_i$ for each class $c_i$.  In
particular,  
\[ TR_1 = \{q_0,q_1,q_2,q_7,q_8\},\]
and
\[ TR_2 = \{q_3,q_4,q_5,q_6,q_9\}. \]
How can we build a classifier for $c_1$ and $c_2$ using models from
$M_{0.5,A}$?
First, we evaluate each model $m$ by how well it has captured the
members of each class.  Define ratings $r_i$ ($i=1,2$) for each $m$ as
\[ r_i(m) = Prob_F (q \in m | q \in TR_i). \]
For example, consider model $m_1 = \{q_3,q_5,q_6,q_8,q_9\}$, where $q_8$ is
in $TR_1$ and the rest are in $TR_2$.  $TR_1$ has 5 members and 1 is
in $m_1$, therefore $r_1(m_1) = 1/5 = 0.2$.  $TR_2$ has (incidentally,
also) 5 members and 4 of them are in $m_1$, therefore $r_2(m_1) = 4/5
= 0.8$.  Thus these ratings represent the quality of the models as a 
description of each class.  A model with a rating $1.0$ for a class is
a perfect model for that class.  We call the difference between $r_1$
and $r_2$ the {\em degree of enrichment} of $m$ with respect to classes
$(1,2)$, i.e., $d_{12} = r_1 - r_2$.  A model $m$ is {\em enriched} if
$d_{12} \neq 0$.  Now we define, for all enriched models $m$,

\[ X_{12}(q,m) = \frac{C_m(q) - r_2(m)}{r_1(m) - r_2(m)}, \]

\noindent and let $X_{12}(q,m)$ be $0$ if $d_{12}(m) = 0$.
For a given $m$, $r_1$ and $r_2$ are fixed, and the value of $X(q,m)$
for each $q$ in $A$ can have one of two values depending on whether
$q$ is in $m$.  For example, for $m_1$, $r_1 = 0.2$ and $r_2 = 0.8$,
so $X(q,m) = -1/3$ for points $q_3,q_5,q_6,q_8,q_9$, and 
$X(q,m) = 4/3$ for points $q_0,q_1,q_2,q_4,q_7$.
Next, for each set $M_t = \{m_1,m_2,...,m_t\}$, we define a
discriminant 

\[ Y_{12}(q,M_t) = \frac{1}{t} \sum_{k=1}^{t} X_{12}(q,m_k). \]

As the set $M_t$ expands,  the value of $Y_{12}$ changes for each $q$.
We show, 
in Table \addtocounter{table}{+3}\thetable\addtocounter{table}{-3}
in the Appendix, the values of $Y_{12}$ for each $M_t$ and each $q$,
and for each new member $m_t$ of $M_t$, $r_1,r_2$, and the two values
of $X_{12}$.  The values of $Y_{12}$ for each $q$ are plotted in
Figure \addtocounter{figure}{1}\thefigure\addtocounter{figure}{-1}.

\begin{figure}[h]
\centering
\begin{tabular}{c}
\psfig{file=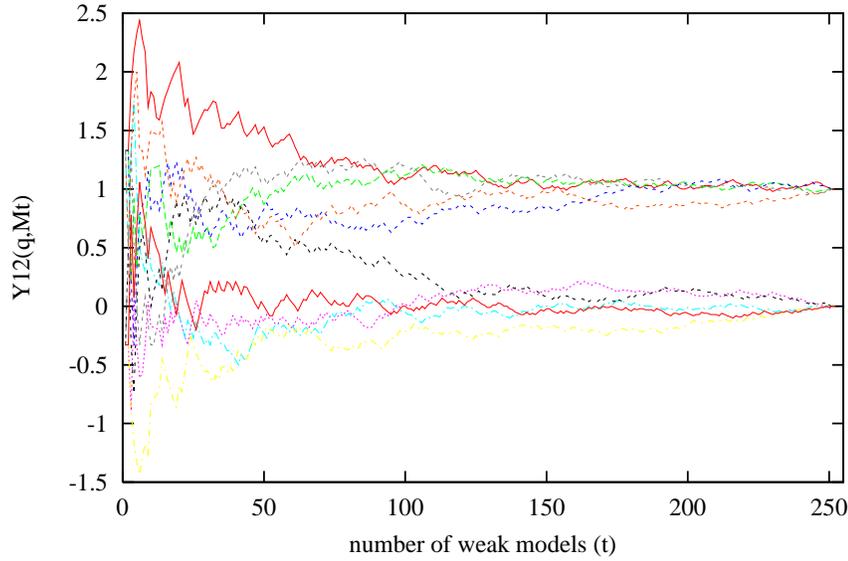,height=3in,clip=}\\
\end{tabular}
\caption{Plot of $Y_{12}(q,M_t)$ versus $t$.  Each line represents the
trace of $Y_{12}(q,M_t)$ for a particular $q$ as $M_t$ expands.}
\end{figure}

In Figure \thefigure \, we see two separate trends.  All those
points that belong to class $c_1$ have their $Y_{12}$ values
converging to 1.0, and all those in $c_2$ converging to 0.0.  Thus
$Y_{12}$ can be used with a threshold 
to classify an arbitrary point $q$.  We can assign $q$ to class $c_1$
if $Y_{12}(q,M_t) > 0.5$,  and to class $c_2$ if $Y_{12}(q, M_t) <
0.5$, and remain undecided when $Y_{12}(q,M_t) = 0.5$.  Observe that
this classifier is fairly accurate far before $M_t$ has expanded to
the full set $M_{0.5,A}$.  We can also change the two poles of $Y_{12}$ to 1.0
and -1.0 respectively by simply rescaling and shifting $X_{12}$:

\[ X_{12}(q,m) = 2(\frac{C_m(q) - r_2(m)}{r_1(m) - r_2(m)}) - 1. \]

How did this separation of trends happen?  Let us now take a closer look at the
models in each $M_t$ and see how many of them cover each point $q$.
For a given $M_t$, among its members, there can be different values of
$r_1$ and $r_2$.  But because of our choices of the
sizes of $TR_1$, $TR_2$, and $m$,  we have only a small set of
distinct values that $r_1$ and $r_2$ can have.  Namely, since each
model has 5 points, there are only six possibilities as follows. 

\begin{center}
\begin{tabular}{l cccccc}
no. of points from $TR_1$ & 0   & 1   & 2   & 3   & 4   & 5   \\
no. of points from $TR_2$ & 5   & 4   & 3   & 2   & 1   & 0   \\
$r_1$                     & 0.0 & 0.2 & 0.4 & 0.6 & 0.8 & 1.0 \\
$r_2$                     & 1.0 & 0.8 & 0.6 & 0.4 & 0.2 & 0.0 \\
\end{tabular}
\end{center}

Note that in a general setting $r_1$ and $r_2$ do not have to sum up
to 1.  If we included models of a larger size, say, one with 10
points,  we can have both $r_1$ and $r_2$ equal to 1.0.  We have
simplified matters by using models of a fixed size and training sets
of the same size.  According to the values of $r_1$ and $r_2$, in this
case we have only 6 different kinds of models.

Now we take a detailed look at the coverage of each point $q$ by
each kind of models, i.e., models of a particular rating (quality) for
each class.   Let us count how many of the models of each value of
$r_1$ and $r_2$ cover each point $q$, and call this $N_{M_t,r_1,TR_1}(q)$ and
$N_{M_t,r_2,TR_2}(q)$ respectively.  We can normalize this count by 
the number of models having each value of $r_1$ or $r_2$,
and obtain a ratio $f_{M_t,r_1,TR_1}(q)$ and 
$f_{M_t,r_2,TR_2}(q)$ respectively.   Thus, for each point $q$, we
have ``a profile of coverage'' by models of each value of ratings
$r_1$ and $r_2$ that is described by these ratios. For example, point $q_0$ at
$t=10$ is only covered by 5 models ($m_2,m_3,m_5,m_8,m_{10}$) in $M_{10}$, and
from Table 3 we know that $M_{10}$ has various numbers of models in
each rating as summarized in the following table.

\begin{center}
\begin{tabular}{l r r r r r r}
$r_1 $                               & 0.0 & 0.2 & 0.4 & 0.6 & 0.8 & 1.0 \\
no. of models in $M_{10}$ with $r_1$ &   0 &   2 &   2 &   4 &   2 &  0 \\
$N_{M_{10},r_1,TR_1}(q_0)$           &   0 &   0 &   0 &   3 &   2 &  0 \\
$f_{M_{10},r_1,TR_1}(q_0)$           &   0 &   0 &   0 & 0.75 & 1.0 & 0 \\
\\
$r_2 $                               & 0.0 & 0.2 & 0.4 & 0.6 & 0.8 & 1.0 \\
no. of models in $M_{10}$ with $r_2$ &   0 &   2 &   4 &   2 &   2 & 0 \\
$N_{M_{10},r_2,TR_2}(q_0)$           &   0 &   2 &   3 &   0 &   0 & 0\\
$f_{M_{10},r_2,TR_2}(q_0)$           &   0 &  1.0 & 0.75 & 0 &   0 & 0\\
\end{tabular}
\end{center}

We show such profiles for each point $q$ and each set $M_t$ in
Figure \addtocounter{figure}{1}\thefigure\addtocounter{figure}{-1}
(as a function of $r_1$) and
Figure \addtocounter{figure}{2}\thefigure\addtocounter{figure}{-2}
(as a function of $r_2$) respectively.

Observe that as $t$ increases,  the profiles of coverage for each point $q$
converge to two distinct patterns.  In Figure 3, the profiles for
points in $TR_1$ converge to a diagonal $f_{M_t,r_1,TR_1} = r_1$, and in
Figure 4, those for points in $TR_2$ also converge to a diagonal
$f_{M_t,r_2,TR_2} = r_2$.  That is, when $M_t = M_{0.5,A}$,  we have
for all $q$ in $TR_1$ and for all $r_1$,
$Prob_{\cal M} (q \in m | m \in M_{r_1,TR_1}) = r_1$, and 
for all $q$ in $TR_2$ and for all $r_2$,
$Prob_{\cal M} (q \in m | m \in M_{r_2,TR_2}) = r_2$.
Thus we have the symmetry in place for both $TR_1$ and $TR_2$.
This is a consequence of $M_t$ being both $TR_1$-uniform and $TR_2$-uniform. 

The discriminant $Y_{12}(q,M_t)$ is a summation over all models $m$ in
$M_t$, which can be decomposed into the sums of terms corresponding to
different ratings $r_i$ for either $i=1$ or $i=2$.  To understand what
happens with the points in $TR_1$, we can decompose their $Y_{12}$ by
values of $r_1$.  Assume that there are $t_x$ models in $M_t$ that
have $r_1 = x$.  Since we have only 6 distinct values for $x$,
$M_t$ is a union of 6 disjoint sets,  and $Y_{12}$ can be decomposed as

\vspace{0.1in}
\begin{tabular}{lll}
\hspace{-0.3in}
$Y_{12}(q,M_t)$ = & $\frac{t_{0.0}}{t} [\frac{1}{t_{0.0}}
\sum_{k_{0.0}=1}^{t_{0.0}}  X_{12}(q,m_{k_{0.0}})] ~~+ $
                  & $\frac{t_{0.2}}{t} [\frac{1}{t_{0.2}}
\sum_{k_{0.2}=1}^{t_{0.2}}  X_{12}(q,m_{k_{0.2}})] ~~+ $\\
                  & $\frac{t_{0.4}}{t} [\frac{1}{t_{0.4}}
\sum_{k_{0.4}=1}^{t_{0.4}}  X_{12}(q,m_{k_{0.4}})] ~~+ $
                  & $\frac{t_{0.6}}{t} [\frac{1}{t_{0.6}}
\sum_{k_{0.6}=1}^{t_{0.6}}  X_{12}(q,m_{k_{0.6}})] ~~+ $\\
                  & $\frac{t_{0.8}}{t} [\frac{1}{t_{0.8}}
\sum_{k_{0.8}=1}^{t_{0.8}}  X_{12}(q,m_{k_{0.8}})] ~~+ $
                  & $\frac{t_{1.0}}{t} [\frac{1}{t_{1.0}}
\sum_{k_{1.0}=1}^{t_{1.0}}  X_{12}(q,m_{k_{1.0}})]$. \\
\end{tabular}
\vspace{0.1in}

The factor in the square bracket of each term is the expectation of values of
$X_{12}$ corresponding to that particular rating $r_1 = x$.  Since
$r_1$ is the same for all $m$ contributing to that term,  by our
choice of sizes of $TR_1$, $TR_2$, and the models, $r_2$ is also the
same for all those $m$ relevant to that term.  Let that value of $r_2$
be $y$, we have, for each (fixed) $q$, each value of $x$ and the
associated value $y$, 

\[ E(X_{12}(q,m_x)) = E(\frac{C_{m_x}(q) - y}{x - y}) = 
\frac{E(C_{m_x}(q))-y}{x-y} = \frac{x-y}{x-y} = 1. \]

The second to the last equality is a consequence of the uniformity of
$M_t$:  because the collection $M_t$ (when $t=252$) covers $TR_1$
uniformly,  we have for each value $x$, 
$Prob_{\cal M} (q \in m | m \in M_{x,TR_1}) = x$, and
since $C_{m_x}(q)$ has only two values (0 or 1), and
$C_{m_x}(q) = 1 ~{\rm iff}~ q \in m$, we have the expected value of 
$C_{m_x}(q)$ equal to $x$.  Therefore 

\[ Y_{12}(q,M_t) = \frac{t_{0.0} + t_{0.2} + t_{0.4} + t_{0.6} +
t_{0.8} + t_{1.0}}{t} = 1. \]

In a more general case,  the values of $r_2$ are not
necessarily equal for all models with the same value for $r_1$, so
we cannot take $y$ and $x-y$ out as constants.  But
then we can further split the term by the values of $r_2$, and proceed
with the same argument.

A similar decomposition of $Y_{12}$ into terms corresponding to
different values of $r_2$ will show that $Y_{12}(q,M_t) = 0$ for those
points in $TR_2$.

\section{Projectability of Models}
We have built a classifier and shown that it works for $TR_1$
and $TR_2$.  How can this classifier work for an arbitrary point that
is not in $TR_1$ or $TR_2$?  Suppose that the feature space $F$
contains other points $p$ (marked by ``,''), and that each $p$ is
close to some training point $q$ (marked by ``.'') as follows.

\begin{center}
\begin{tabular}{cccccccccc}
.,&.,&.,&.,&.,&.,&.,&.,&.,&.,\\
$q_0,p_0$& $q_1,p_1$& $q_2,p_2$& $q_3,p_3$& $q_4,p_4$&
$q_5,p_5$& $q_6,p_6$& $q_7,p_7$& $q_8,p_8$& $q_9,p_9$\\
\end{tabular}
\end{center}

We can take the models $m$ as regions in the space that cover the
points $q$ in the same manner as before.  Say, if each point $q_i$
has a particular value of the feature $v$ (in our one-dimensional
feature space) that is $v(q_i)$.  We can define a model by ranges of values for
this feature, e.g., in our example $m_1$ covers
$q_3,q_5,q_6,q_8,q_9$, so we take

\begin{center}
\begin{tabular}{ll}
$m_1 = $ & $\{q | \frac{v(q_2)+v(q_3)}{2} < v(q) <
\frac{v(q_3)+v(q_4)}{2}\} \cup $\\
& $\{q | \frac{v_(q_4)+v(q_5)}{2} < v(q) < \frac{v(q_6)+v(q_7)}{2} \} \cup$ \\
& $\{q | \frac{v_(q_7)+v(q_8)}{2} < v(q) \}. $\\
\end{tabular}
\end{center}

Thus we can tell if an arbitrary point $p$ with value $v(p)$ for this
feature is inside or outside this model.

We can calculate the model's ratings in exactly the same way as
before,  using only the points $q$.  But now the same 
classifier works for the new points $p$,  since we can use the new
definitions of models to determine if $p$ is inside or outside each model.
Given the proximity relationship as above,
those points will be assigned to the same class as their closest
neighboring $q$.  If these are indeed the true classes for the points
$p$,  the classifier is perfect for this new set.  In the SD terminology, if we
call the two subsets of points $p$ that should be labeled as two
different classes $TE_1$ and $TE_2$,  i.e., $TE_1 =
\{p_0,p_1,p_2,p_7,p_8\}$, $TE_2 = \{p_3,p_4,p_5,p_6,p_9\}$, we say
that $TR_1$ and $TE_1$ are $M_t$-indiscernible,  and similarly $TR_2$
and $TE_2$ are also $M_t$-indiscernible.   This is to say,  from the
perspective of $M_t$, 
there is no difference between $TR_1$ and $TE_1$,  or $TR_2$ and
$TE_2$, therefore all the properties of $M_t$ that are observed using
$TR_1$ and $TR_2$ can be projected to $TE_1$ and $TE_2$.
The central challenge of an SD method is to maintain projectability,
uniformity, and enrichment of the collection of models at the same
time. 

\section{Developments of SD Theory and Algorithms}

\subsection{Algorithmic Implementations}

The method of stochastic discrimination constructs a classifier by
combining a large number of simple discriminators that are called {\em
weak models}.   A weak model is simply a subset of the feature space.
Conceptually,  the classifier is constructed by a three-step process:
(1) weak model generation, (2) weak model evaluation, and (3) weak
model combination.   The generator enumerates weak models in an
arbitrary order and passes them on to the evaluator.   The evaluator
has access to the training set.   It rates and filters the weak models
according to their capability in capturing points of each class,
and their contribution to satisfying the uniformity condition.
The combinator then produces a discriminant function that depends on
a point's membership status with respect to each model, and the
models' ratings.  At classification,  a point is assigned to the class
for which this discriminant has the highest value.  Informally, the 
method captures the intuition of gaining wisdom from graded random
guesses.  

\subsubsection*{Weak model generation.}

Two guidelines should be observed in generating the weak models:

(1) {\em projectability}:   A weak model should be able to capture more than
one point so that the solution can be projectable to points not
included in the training set.
Geometrically,  this means that a useful model must be of
certain minimum size,  and it should be able to capture points that
are considered {\em neighbors} of one another.
To guarantee similar accuracies of the classifier (based on
similar ratings of the weak models) on both training and testing data,  
one also needs an assumption that the training data are {\em representative}. 
Data representativeness and model projectability are two sides of
the same question.  More discussions of this can be found in
\cite{kn:berlind94}.
A weak model defines a {\em neighborhood} in the space,  and
we need a training sample in a neighborhood of every unseen sample.
Otherwise,  since our only knowledge of the class boundaries is from the given
training set,  there can be no basis for any inference concerning
regions of the feature space where no training samples are given.

(2) {\em simplicity of representation}:  A weak model should have a
simple representation.   That means, the membership of an
arbitrary point with respect to a model must be cheaply computable.
To illustrate this, consider representing a model as a listing of all
the points it contains.  This is practically useless since the
resultant solution could be as expensive as an exhaustive template
matching using all the points in the feature space.   An example of a
model with a simple representation is a half-plane in a
two-dimensional feature space.

Conditions (1) and (2) restrict the type of weak models yet by no
means reduce the number of candidates to any tangible limit.
To obtain an unbiased collection of the candidates with minimum effort,
random sampling with replacement is useful.
The training of the method thus relies on a stochastic process
which, at each iteration, generates a weak model that satisfies the
above conditions.

A convenient way to generate weak models randomly is to use a type of
model that can be described by a small number of parameters.
The values of the parameters can be chosen pseudo-randomly.
Some example types of models that can be generated this way include
(1) half-spaces bounded by a threshold on a randomly selected feature
dimension; 
(2) half-spaces bounded by a hyperplane of equi-distance to two
randomly selected points;
(3) regions bounded by two parallel hyperplanes perpendicular to a
randomly selected axis. 
(4) hypercubes centered at randomly selected points with edges of
varying lengths; 
(5) balls (based on the city-block metric) centered at randomly
selected points with randomly selected radii; and
(6) balls (based on the Euclidean metric) centered at a randomly
selected points with randomly selected radii.
A model can also be a union or intersection of several regions of
these types.  An implementation of SD using hyper-rectangular boxes as
weak models is described in \cite{kn:ho96}.

A number of heuristics may be used in creating these models.
These heuristics specify the way random points are chosen from the space,
or set limits on the maximum and minimum sizes of the models.
By this we mean restricting the choice of random points to, for instance,
points in the space whose coordinates fall inside the range of those of the
training samples,  or restricting the radii of the balls
to, for instance, a fraction of the range of values in a particular feature
dimension.  The purpose of these heuristics is to speed up the search
for acceptable models by confining the search within the most
interesting regions,  or to guarantee a minimum model size.

\subsubsection*{Enrichment enforcement.}

The enrichment condition is relatively easy to enforce,  as models
biased towards one class are most common.
But since the strength of the biases ($|d_{ij}(m)|$)
determines the rate at which accuracy increases,  we tend to prefer to
use models with an enrichment degree further away from zero.

One way to implement this is to use a threshold on the enrichment degree
to select weak models from the random stream so that they are of some
minimum quality.  In this way, one will be able to
use a smaller collection of models to yield a classifier of the
same level of accuracy.  However,  there are tradeoffs involved in
doing this.  For one thing,  models of higher rating are less likely
to appear in the stream, and so more random models have to be explored
in order to find sufficient numbers of higher quality weak models.  And
once the type of model is fixed and the value of the threshold is
set,  there is a risk that such models may never be found.

Alternatively,  one can use the most enriched model found in a
pre-determined number of trials.  This also makes the time needed for
training more predictable,  and it permits a tradeoff between training
time and quality of the weak models.

In enriching the model stream, it is important to remember that if the
quality of weak models selected is allowed to get too high, there is a
risk that they will become training set specific, that is, less
likely to be projectable to unseen samples.  This could present a problem
since the projectability of the final classifier is directly based on
the projectability of its component weak models.  

\subsubsection*{Uniformity promotion.}

The uniformity condition is much more difficult to satisfy.
Strict uniformity requires that every point be covered by the same
number of weak models of every combination of per-class ratings.
This is rather infeasible for continuous and unconstrained ratings.

One useful strategy is to use only weak models of a particular rating.
In such cases, the ratings $r_i(m)$ and $r_j(m)$ are the same for all
models $m$ enriched for the discrimination between classes $i$ and
$j$,  so we need only to make sure that each point is included in the
same number of models.  To enforce this,  models can be created in
groups such that each group partitions the entire space into a set of
non-overlapping regions.  An example is to use leaves of a fully-split
decision tree,  where each leave is perfectly enriched for one class,
and each point is covered by exactly one leave of each tree.  For any pairwise
discrimination between classes $i$ and $j$,  we can use only those
leaves of the trees that contain only points of class $i$.  In other
words,  $r_i(m)$ is always 1 and $r_j(m)$ is always 0.
Constraints are put in the tree-construction process to guarantee
some minimum projectability.

With other types of models,  a first step to promote uniformity is
to use models that are unions of small regions with simple boundaries.
The component regions may be scattered throughout the space.
These models have simple representations but can describe complicated
class boundaries.   They can have some minimum size and hence good
projectability.  At the same time,  the scattered locations of
component regions do not tend to cover large areas repeatedly.

A more sophisticated way to promote uniformity involves defining a measure
of the lack of uniformity and an algorithm to minimize such a measure.
The goal is to create or retain more models located in areas where the
coverage is thinner.    An example of such a measure is the count of those
points that are covered by less-than-average number of previously
retained models.   For each point $x$ in the class $c_0$ to be
positively enriched, we calculate, out of all previous models used for
that class, how many of them have covered $x$.
If the coverage is less than the average for class $c_0$,  we call $x$
a weak point.   When a new model is created,  we check how many
such weak points are covered by the new model.   
The ratio of the set of covered weak points to the set of all the weak
points is used as a merit score of how well this model improves
uniformity.   We can accept only those models with a score over a
pre-set threshold,  or take the model with the best score found in
a pre-set number of trials.
One can go further to introduce a bias to the model generator so that
models covering the weak points are more likely to be created.   The
later turns out to be a very effective strategy that led to good
results in our experiments.

\subsection{Alternative Discriminants and Approximate Uniformity}

The method outlined above allows for rich possibilities of variation at
the algorithmic level.  The variations may be in the design of the
weak model generator, or in ways to enforce the enrichment and uniformity
conditions.  It is also possible to change the definition of the
discriminant,  or to use different kinds of ratings.   

A variant of the discriminating function is studied in detail in
\cite{kn:berlind94}.  In this variant,  we define the ratings by

\[ {r'}_i (m) = {  | m \cap TR_i |  \over {|m \cap TR|}  }, \]

\noindent for all $i$.  It is an estimate of the posterior probability that a
point belongs to class $i$ given the condition that it is included in
model $m$.   The discriminant for class $i$ is defined to be:

\[ W_i(q) = {{ \sum_{k=1,...,p_i} C_m(q) {r'}_i(m) } \over 
	{ \sum_{k=1,...,p_i} C_m(q)}}. \]

\noindent where $p_i$ is the number of models accumulated for class $i$.

It turns out that, with this discriminant, the classifier also approaches
perfection asymptotically provided that an additional {\em symmetry}
condition is satisfied.   The symmetry condition requires that the
ensemble includes the same number of models for all permutations of
$(r'_1, r'_2,..., r'_n)$.  It prevents biases created by using more
$(i,j)$-enriched models than $(j,i)$-enriched models for all pairs
$(i,j)$ \cite{kn:berlind94}.   Again, this condition may be enforced
by using only certain particular permutations of the $r'$ ratings
\cite{kn:ho95}.   This alternative discriminant is convenient
for multi-class discrimination problems.

The SD theory established the mathematical concepts of 
enrichment, uniformity, and projectability of a weak model ensemble.
Bounds on classification accuracy are developed based on
strict requirements on these conditions,  which is 
a mathematical idealization.  In practice, there are often
difficult tradeoffs among the three conditions.
Thus it is important to understand how much of the 
classification performance is affected when these conditions
are weakened.  This is the subject of study in \cite{kn:chen98},
where notions of near uniformity and weak indiscernibility
are introduced and their implications are studied.

\subsection{Structured Collections of Weak Models}

As a constructive procedure, the method of stochastic discrimination
depends on a detailed control of the uniformity of model coverage,
which is outlined but not fully published in the
literature \cite{kn:kleinberg00}.  The method of random subspaces 
followed these ideas but attempted a different approach.   Instead of
obtaining weak discrimination and projectability through simplicity of
the model form, and forcing uniformity by sophisticated algorithms,  the
method uses complete, locally pure partitions as given in fully split
decision trees \cite{kn:ho98} or nearest neighbor classifiers \cite{kn:ho98a}
to achieve strong discrimination and uniformity,  and then explicitly
forces different generalization patterns on the component classifiers.
This is done by training large capacity component classifiers such as
nearest neighbors and decision trees to fully fit the data,  but
restricting the training of each classifier to a coordinate subspace
of the feature space where all the data points are projected,  so
that classifications remain invariant in
the complement subspace.  If there is no ambiguity in the subspaces,  the
individual classifiers maintain maximum accuracy on the training data,
with no cases deliberately chosen to be sacrificed,  and thus the
method does not run into the paradox of sacrificing some training
points in the hope for better generalization accuracy.
This is to create a collection of weak models in a structured way.

However the tension among the three factors persists.   There is
another difficult tradeoff in how much discriminating power to retain
for the component classifiers.  Can every one use only a single
feature dimension so as to maximize invariance in the complement
dimensions?   Also,  projection to coordinate subspaces sets parts of
the decision boundaries parallel to the coordinate axes.  Augmenting
the raw features by simple transformations \cite{kn:ho98} introduces
more flexibility, but it may still be insufficient for an arbitrary
problem.  Optimization of generalization performance will continue to
depend on a detailed control of the projections to suit a particular
problem. 

\section{Conclusions}

The theory of stochastic discrimination identifies three and only
three sufficient conditions for a classifier to achieve maximum
accuracy for a problem.   These are just the three elements long
believed to be important in pattern recognition:  discrimination
power, complementary information,  and generalization ability.
It sets a foundation for theories of ensemble learning.
Many current questions on classifier combination can have an answer
in the arguments of the SD theory:
What is good about building the classifier on weak models instead of
strong models?  Because weak models are easier to obtain,  and their
smaller capacity renders them less sensitive to sampling errors in
small training sets \cite{kn:vapnik82} \cite{kn:vapnik98}, thus
they are more likely to have similar coverage on the unseen points
from the same problem.  Why are many models needed?   Because the
method relies on the law of large numbers to reduce the variance of
the discriminant on each single point.  How should these models
complement each other?  The uniformity condition specifies exactly what
kind of correlation is needed among the individual models. 

Finally, we emphasize that the accuracy of SD methods is not achieved
by intentionally limiting the VC dimension of the complete system; the
combination of many weak models can have a very large VC dimension.
It is a consequence of the symmetry relating probabilities in the two
spaces, and the law of large numbers.  It is a structural property of
the topology.   The observation of this symmetry and its relationship
to ensemble learning is a deep insight of Kleinberg's that we believe
can lead to better understanding of other ensemble methods.

\section*{Acknowledgements}
The author thanks Eugene Kleinberg for many discussions over the past
decade on the theory of stochastic discrimination, its comparison
to other approaches, and perspectives on the fundamental issues in
pattern recognition.

\begin{figure}[p]
\centering
\begin{tabular}{cc}
\psfig{file=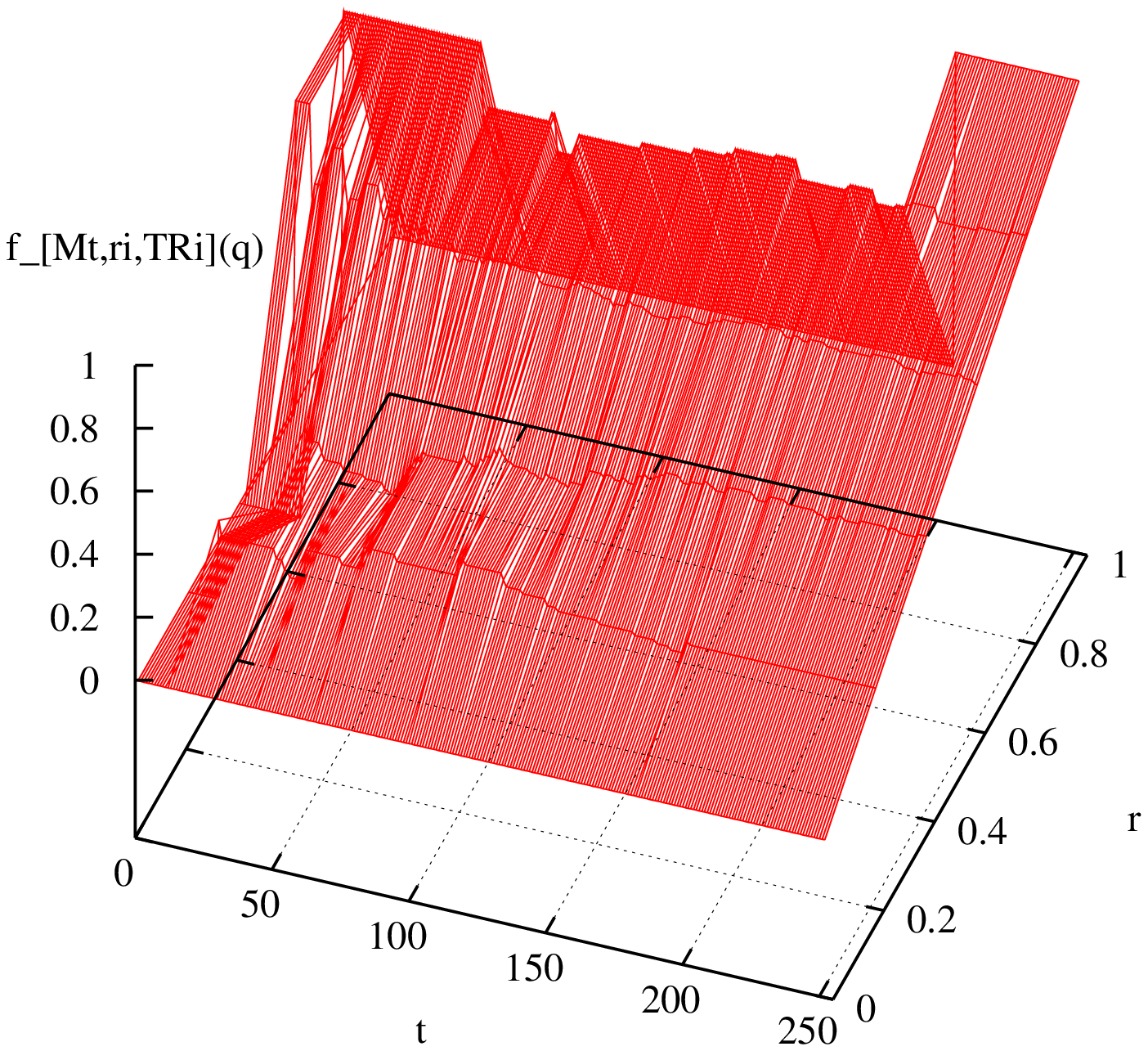,height=1.5in,clip=}&
\psfig{file=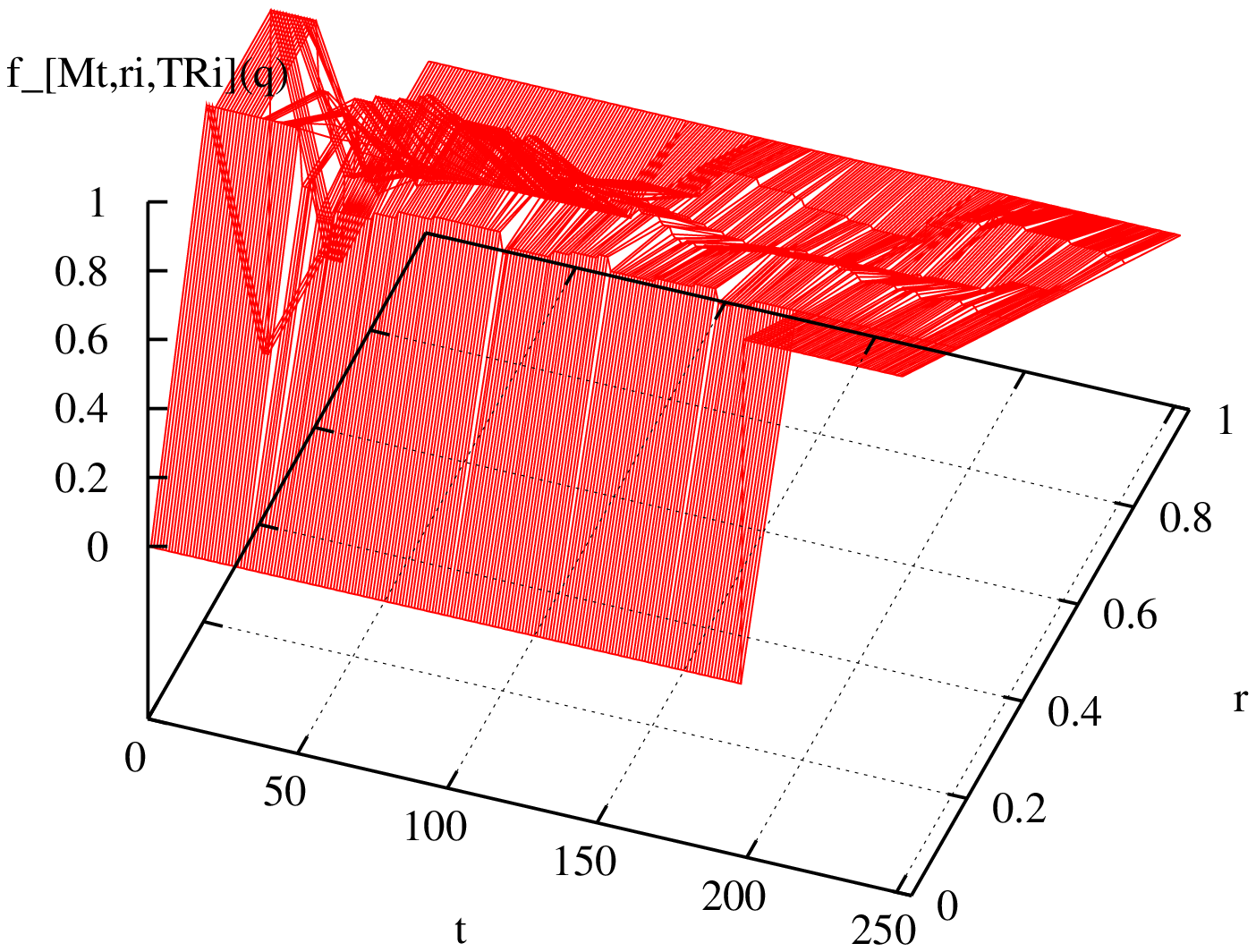,height=1.5in,clip=} \\
$q_0 \in TR_1$ & $q_5 \in TR_2$ \\
\psfig{file=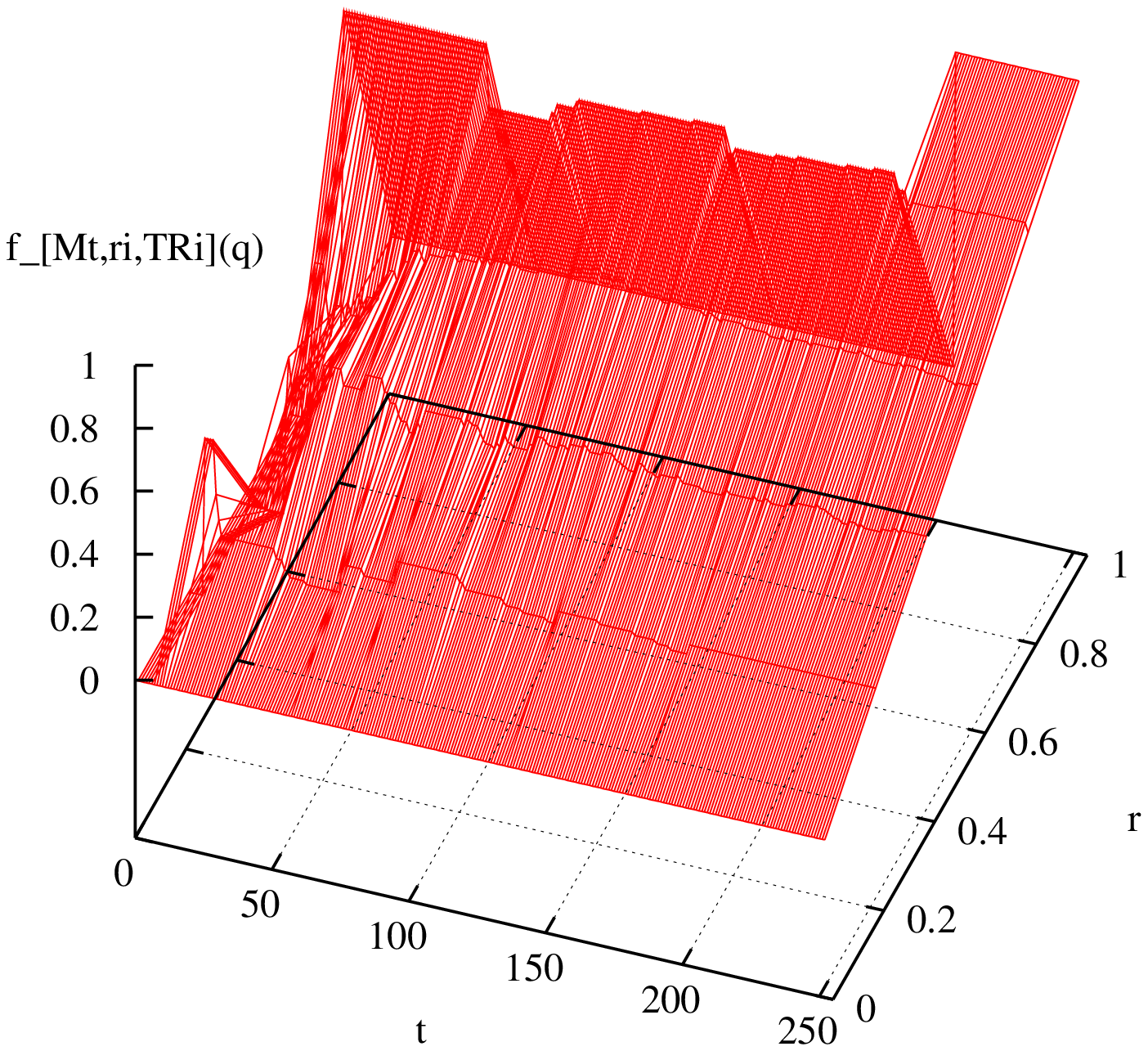,height=1.5in,clip=}&
\psfig{file=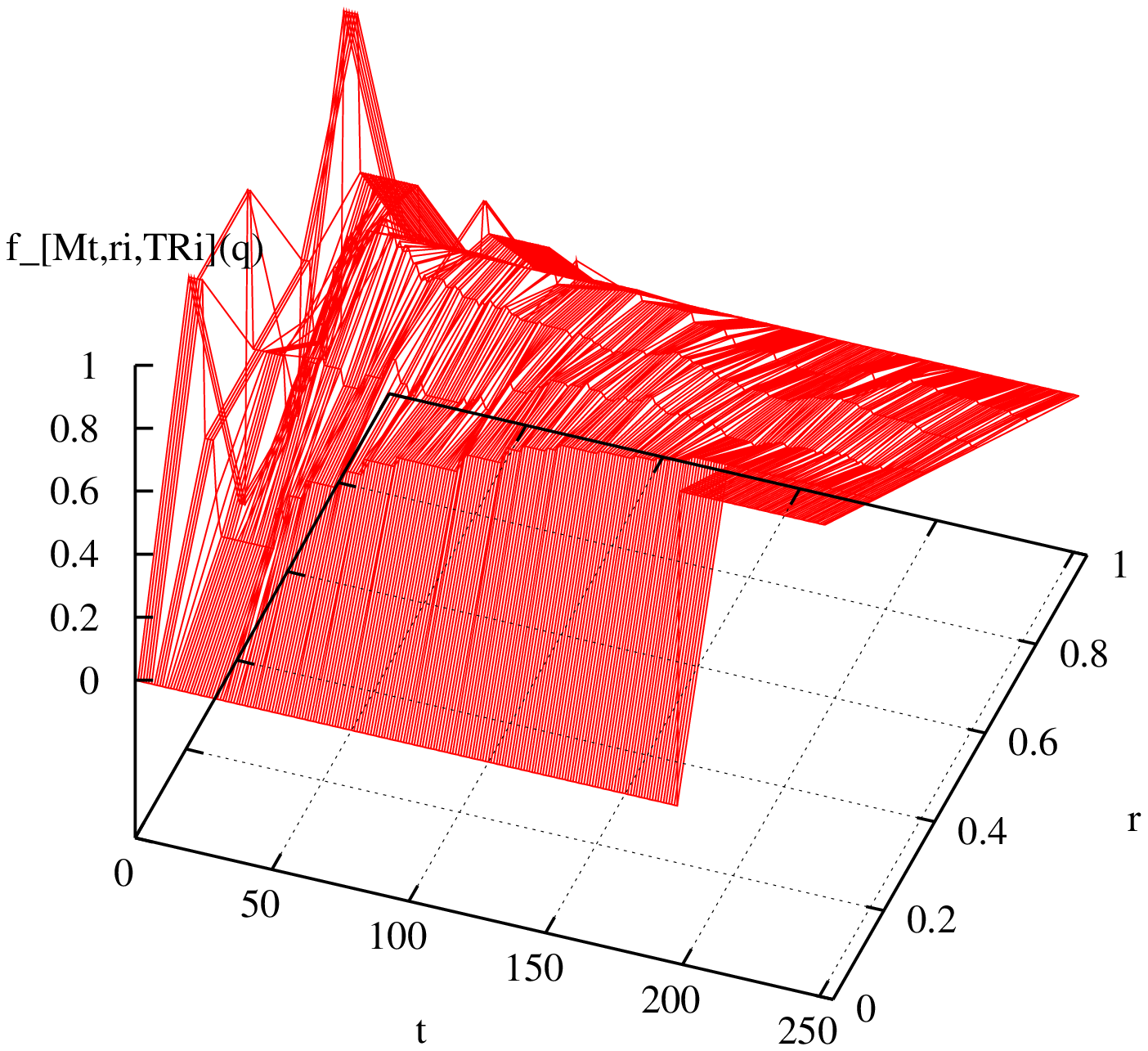,height=1.5in,clip=}\\
$q_1 \in TR_1$ & $q_6 \in TR_2$ \\
\psfig{file=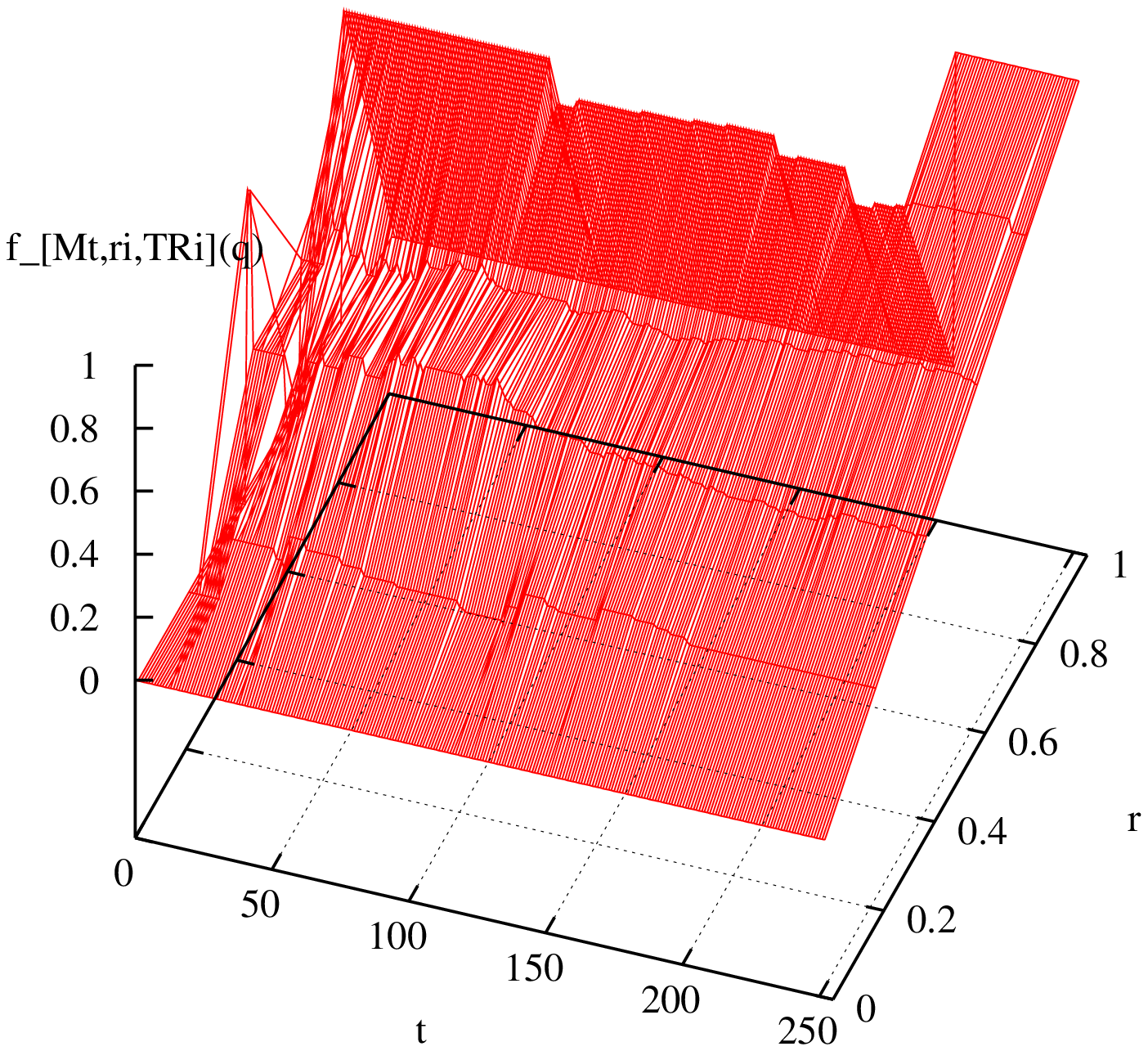,height=1.5in,clip=}&
\psfig{file=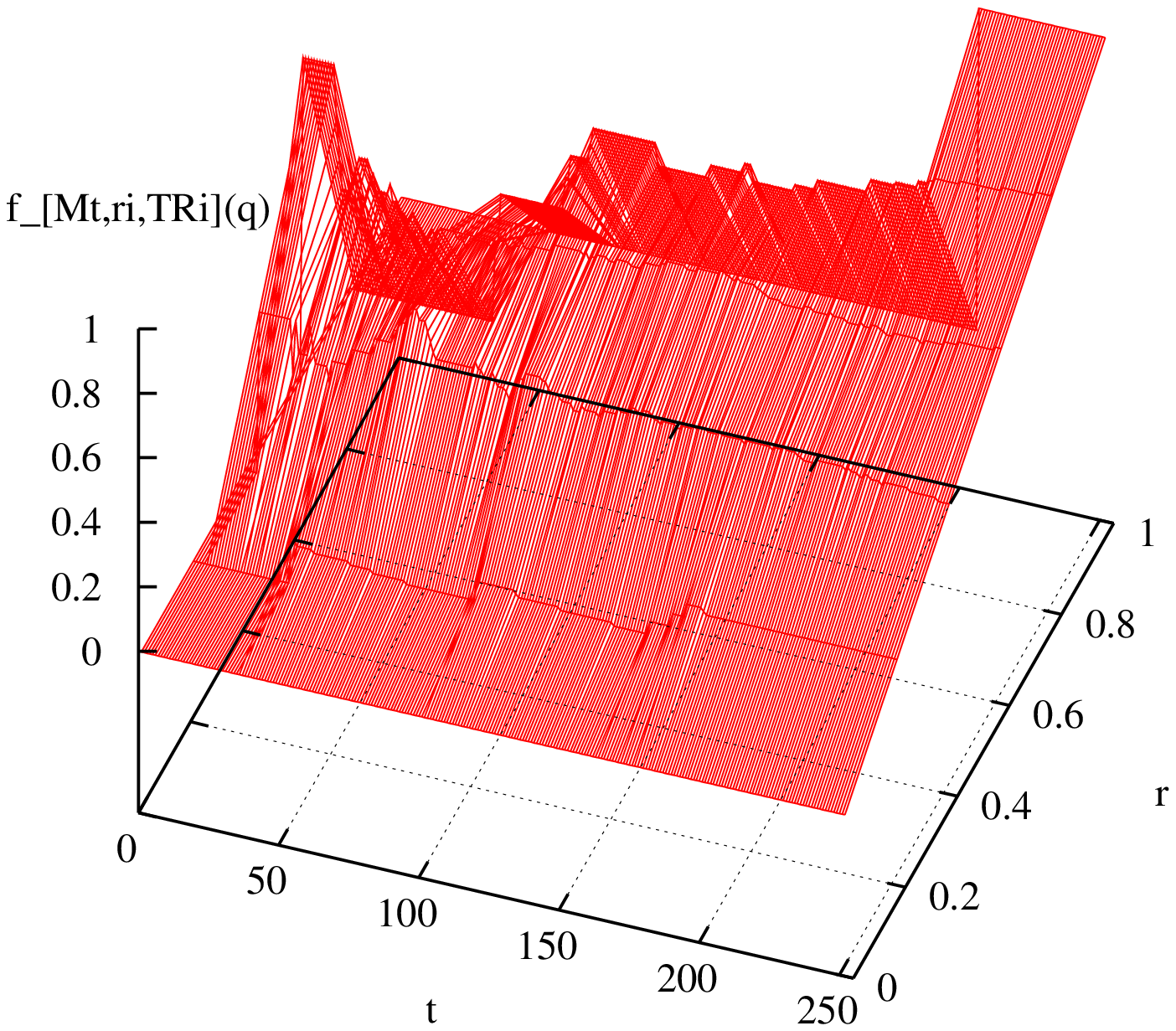,height=1.5in,clip=}\\
$q_2 \in TR_1$ & $q_7 \in TR_1$ \\
\psfig{file=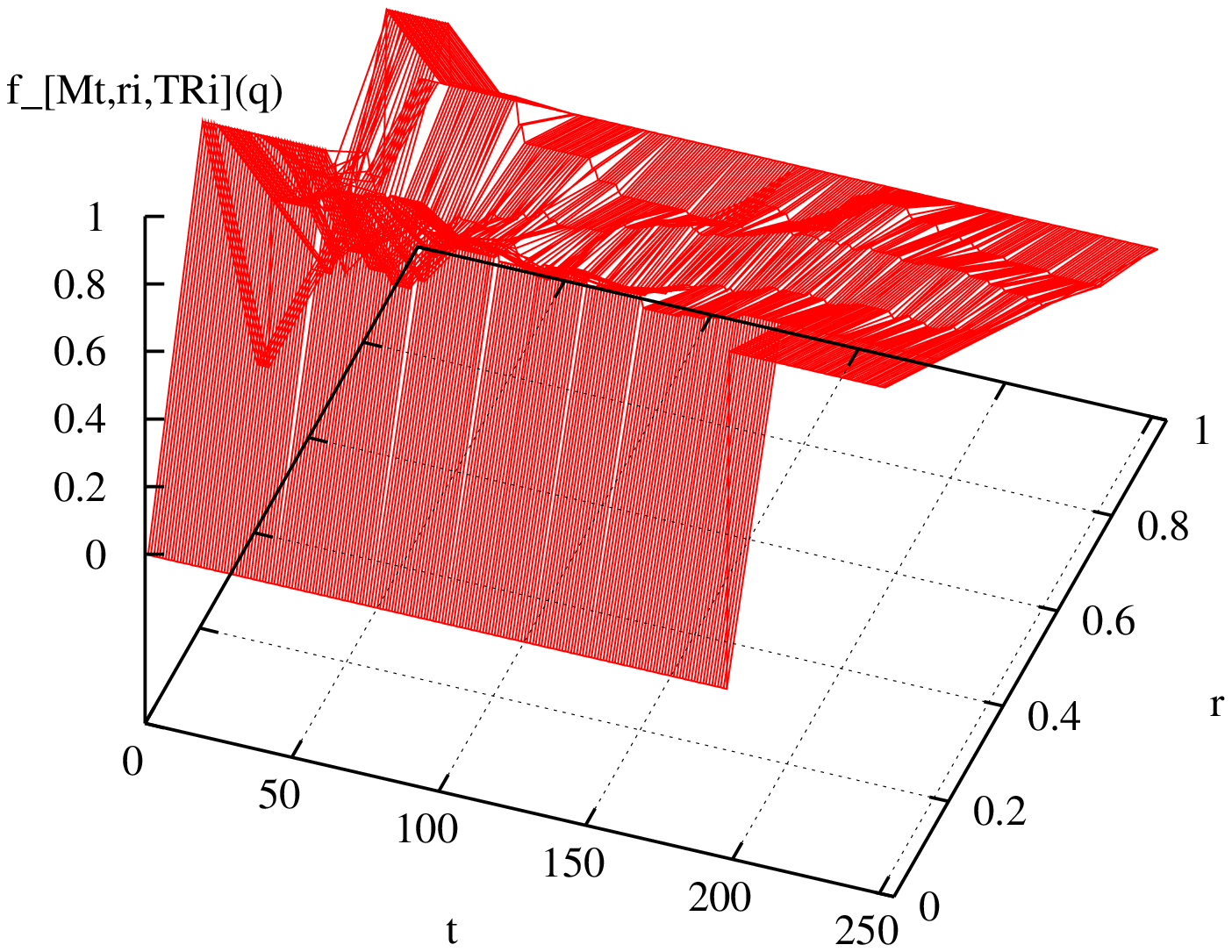,height=1.5in,clip=}&
\psfig{file=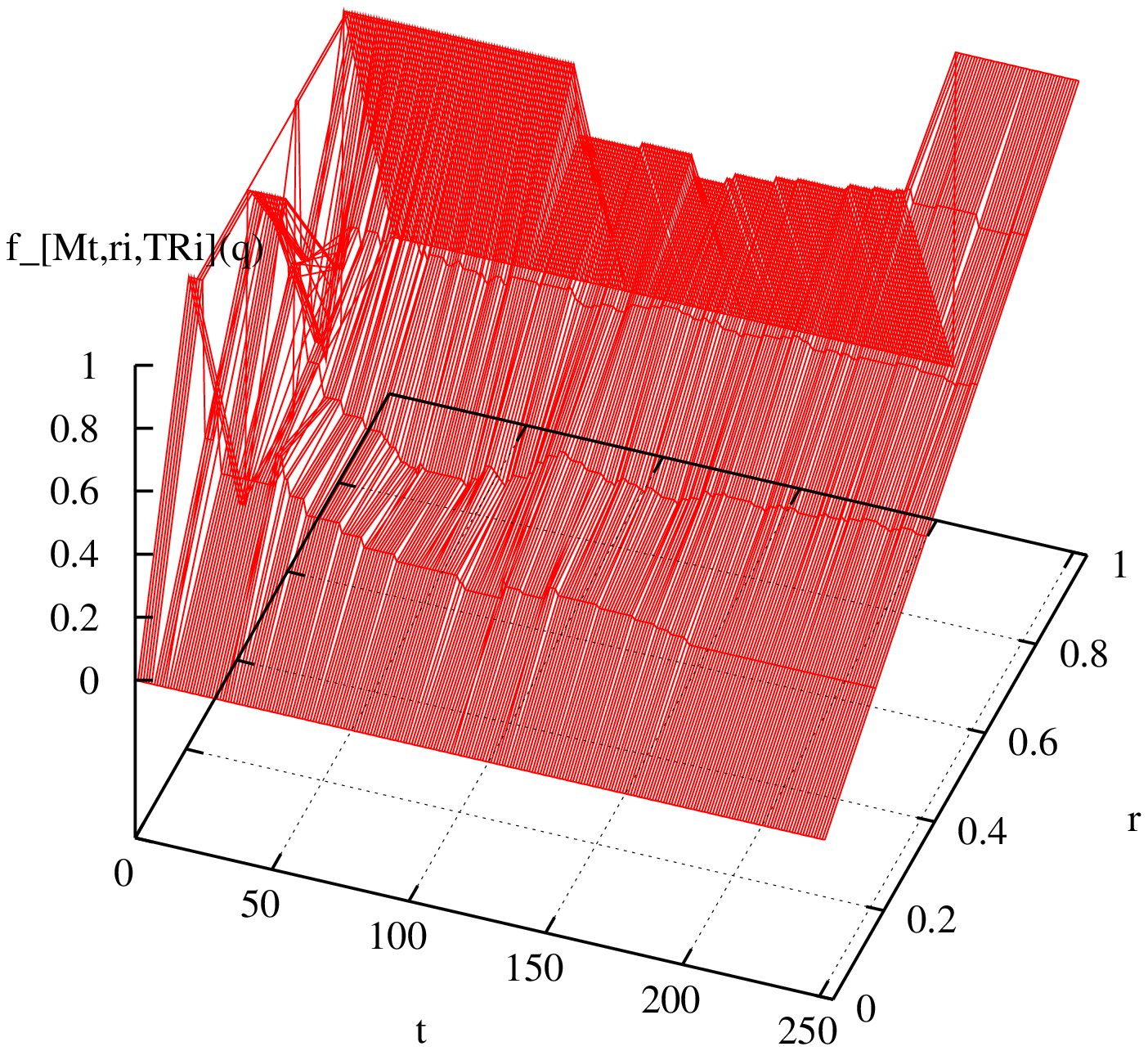,height=1.5in,clip=}\\
$q_3 \in TR_2$ & $q_8 \in TR_1$ \\
\psfig{file=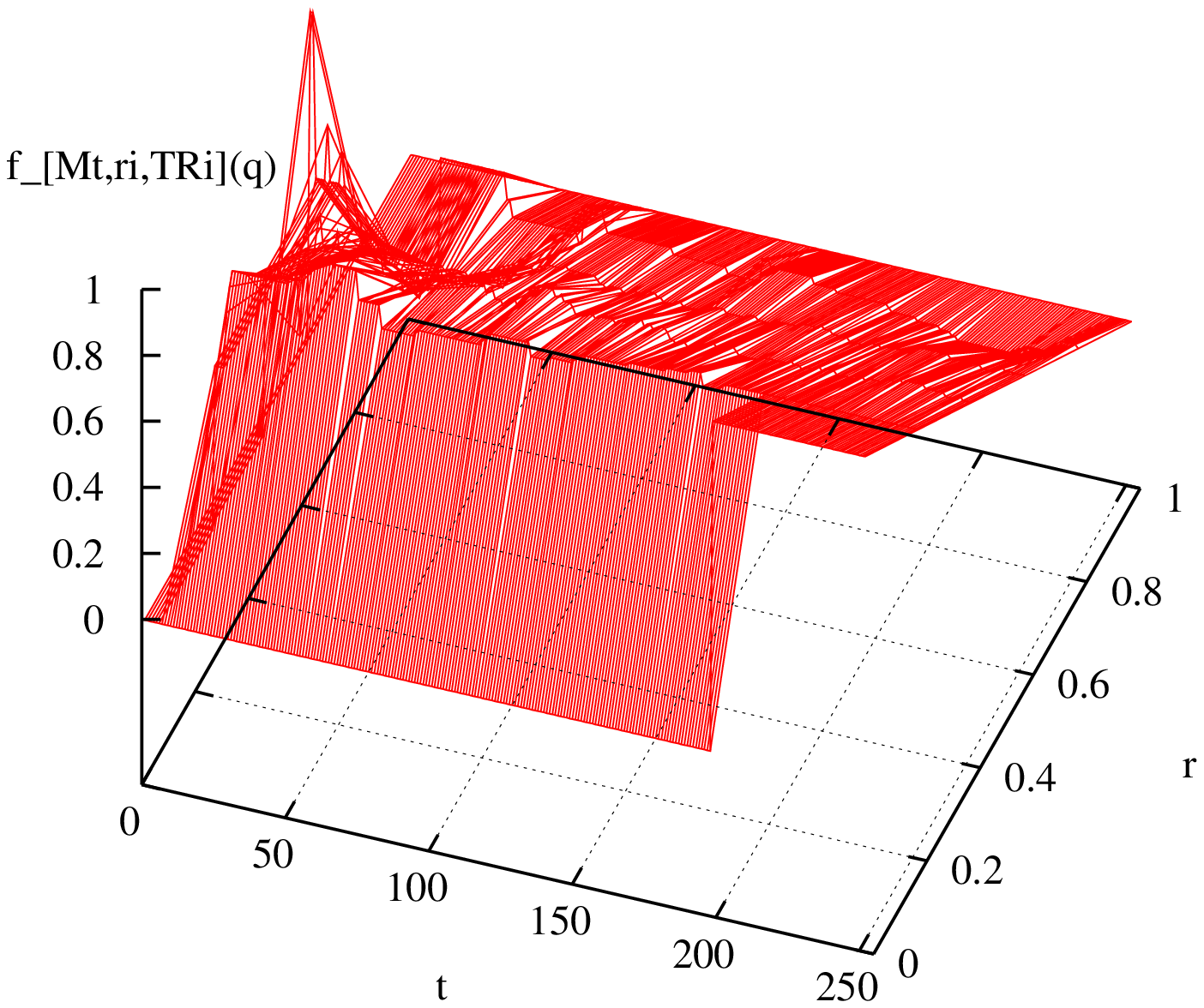,height=1.5in,clip=}&
\psfig{file=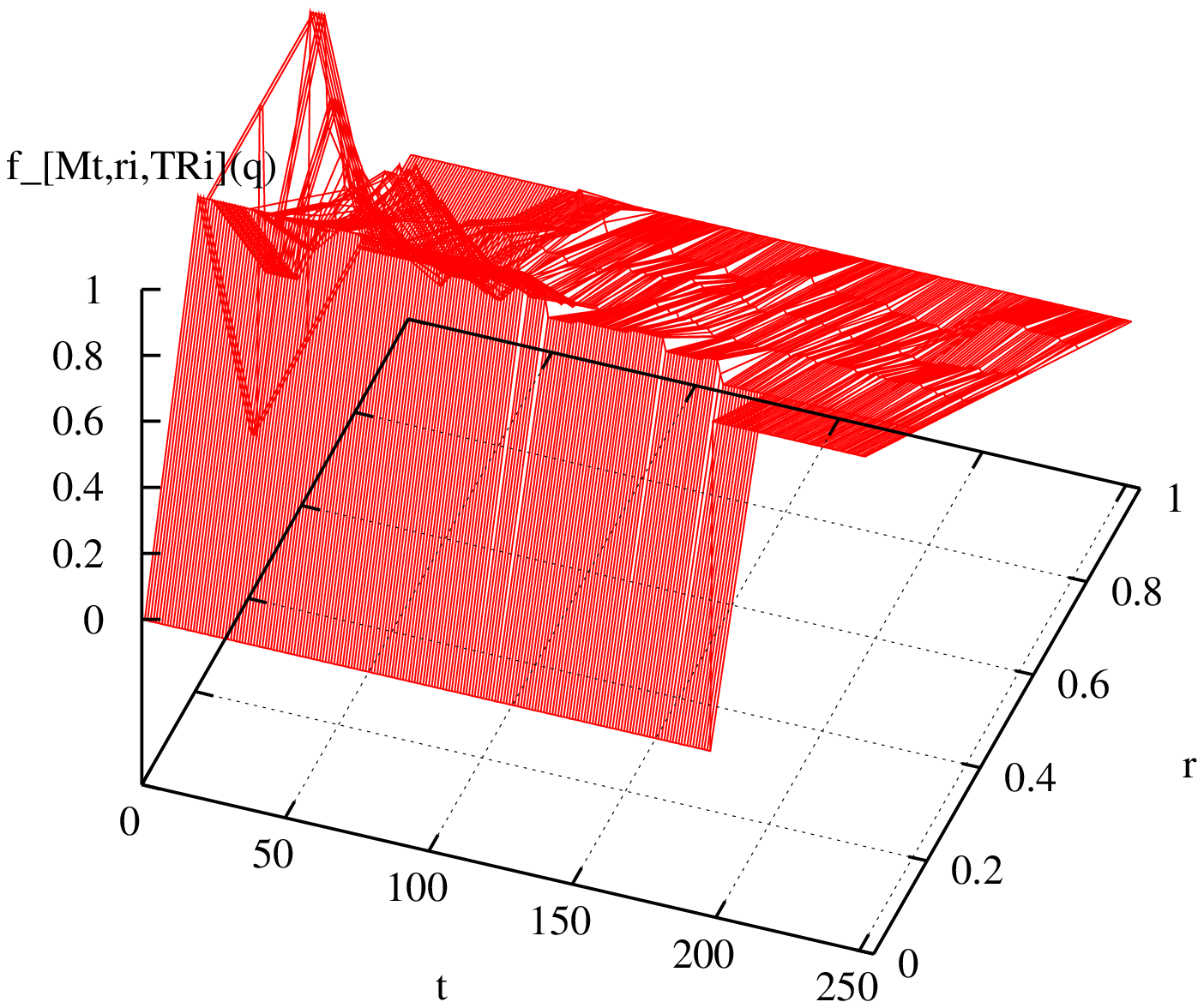,height=1.5in,clip=}\\
$q_4 \in TR_2$ & $q_9 \in TR_2$ \\
\end{tabular}
\caption{$f_{M_t,r_1,TR_1}(q)$ for each point $q$ and set $M_t$.
In each plot, the $x$ axis is $t$ that ranges from 0 to 252, the $y$
axis is $r$ that ranges from $0$ to $1$, and the $z$ axis is 
$f_{M_t,r_1,TR_1}$.}
\end{figure}

\begin{figure}[p]
\centering
\begin{tabular}{cc}
\psfig{file=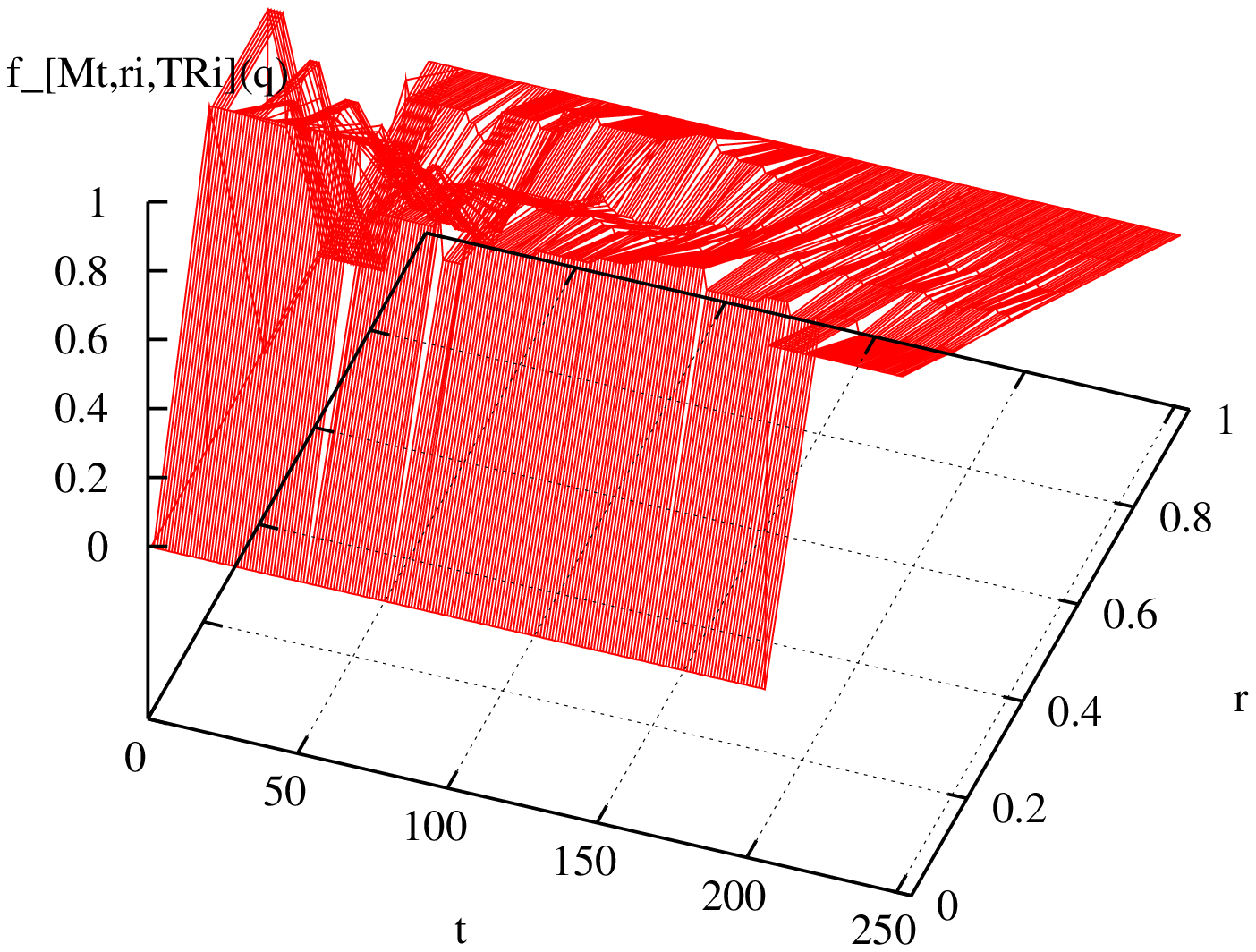,height=1.5in,clip=}&
\psfig{file=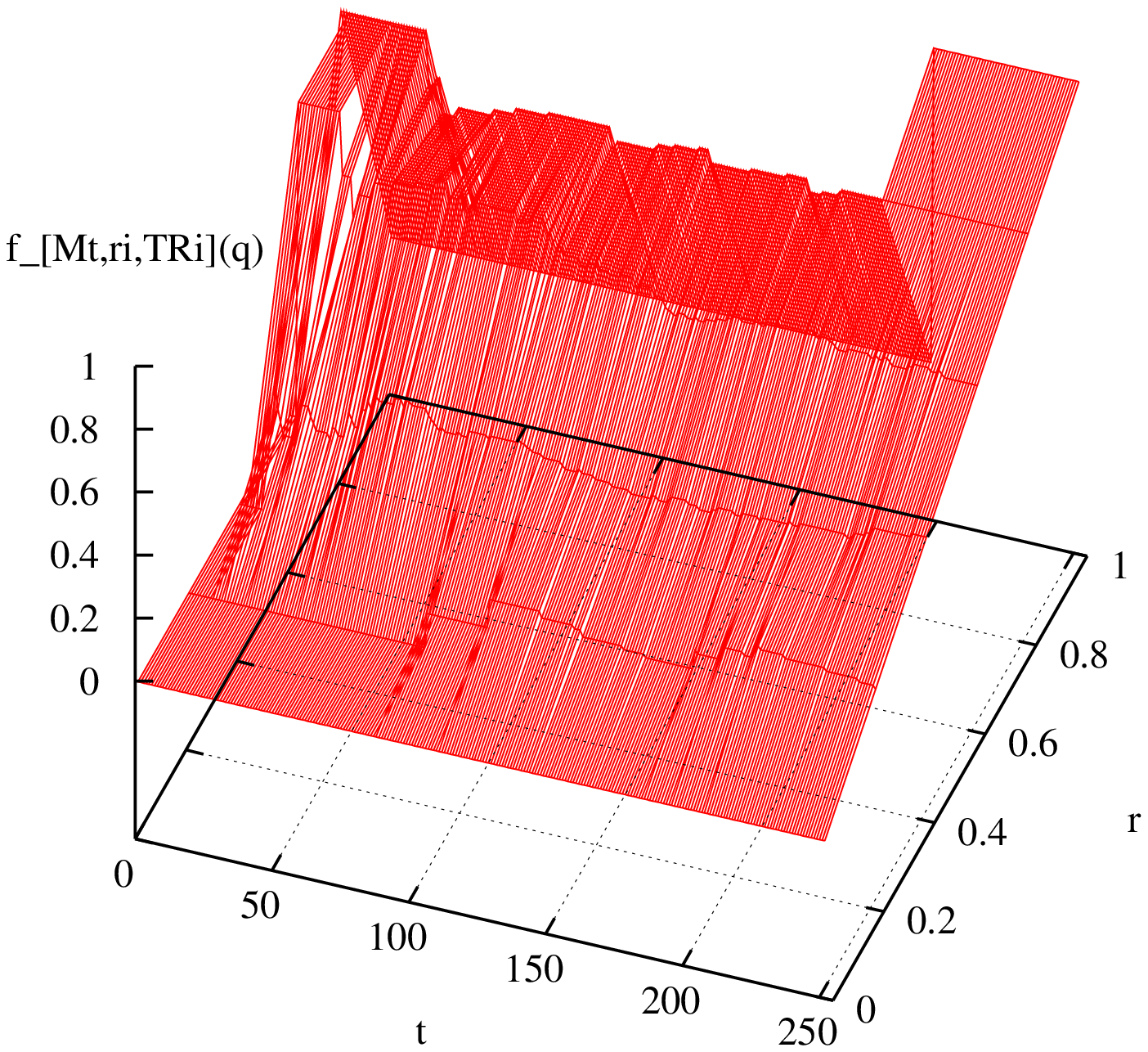,height=1.5in,clip=}\\
$q_0 \in TR_1$ & $q_5 \in TR_2$ \\
\psfig{file=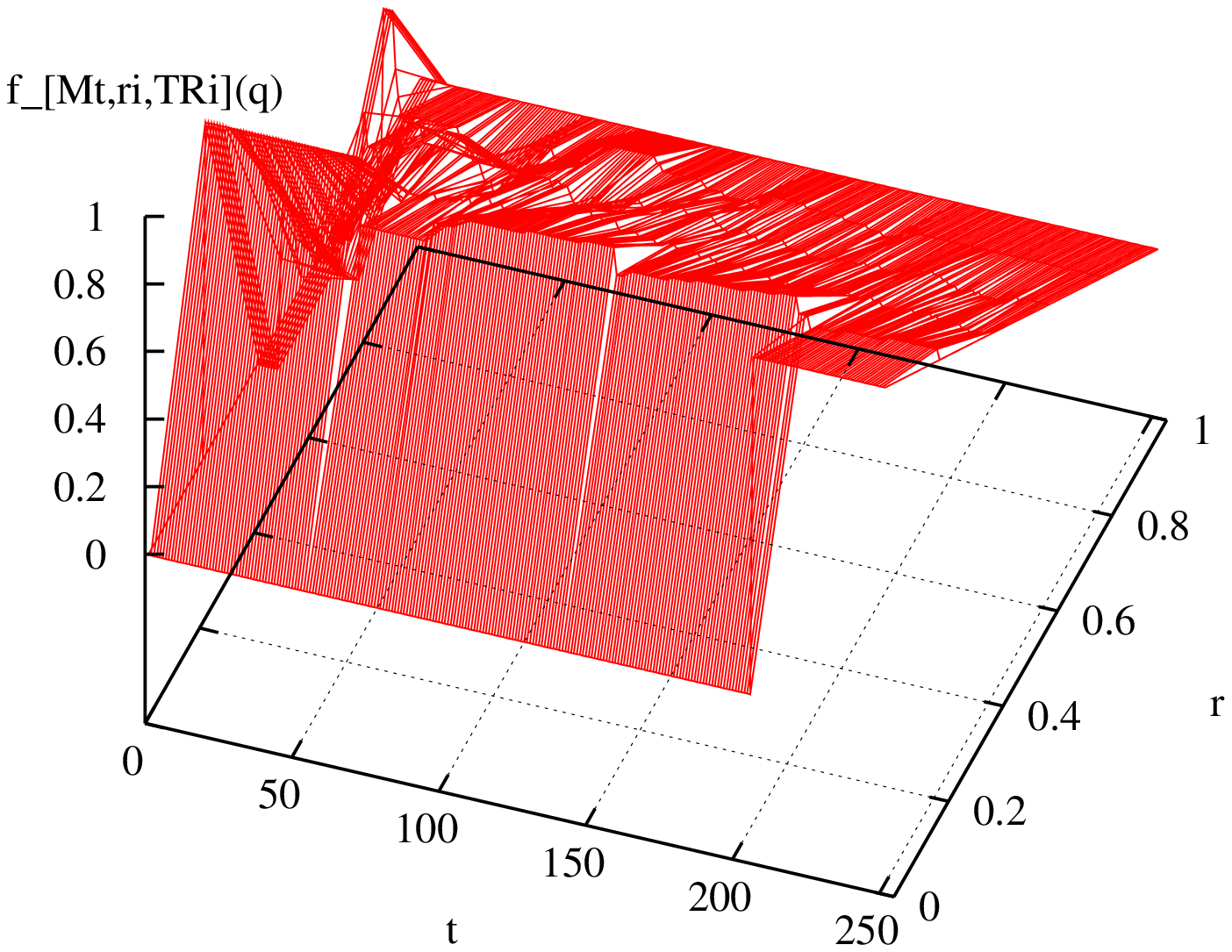,height=1.5in,clip=}&
\psfig{file=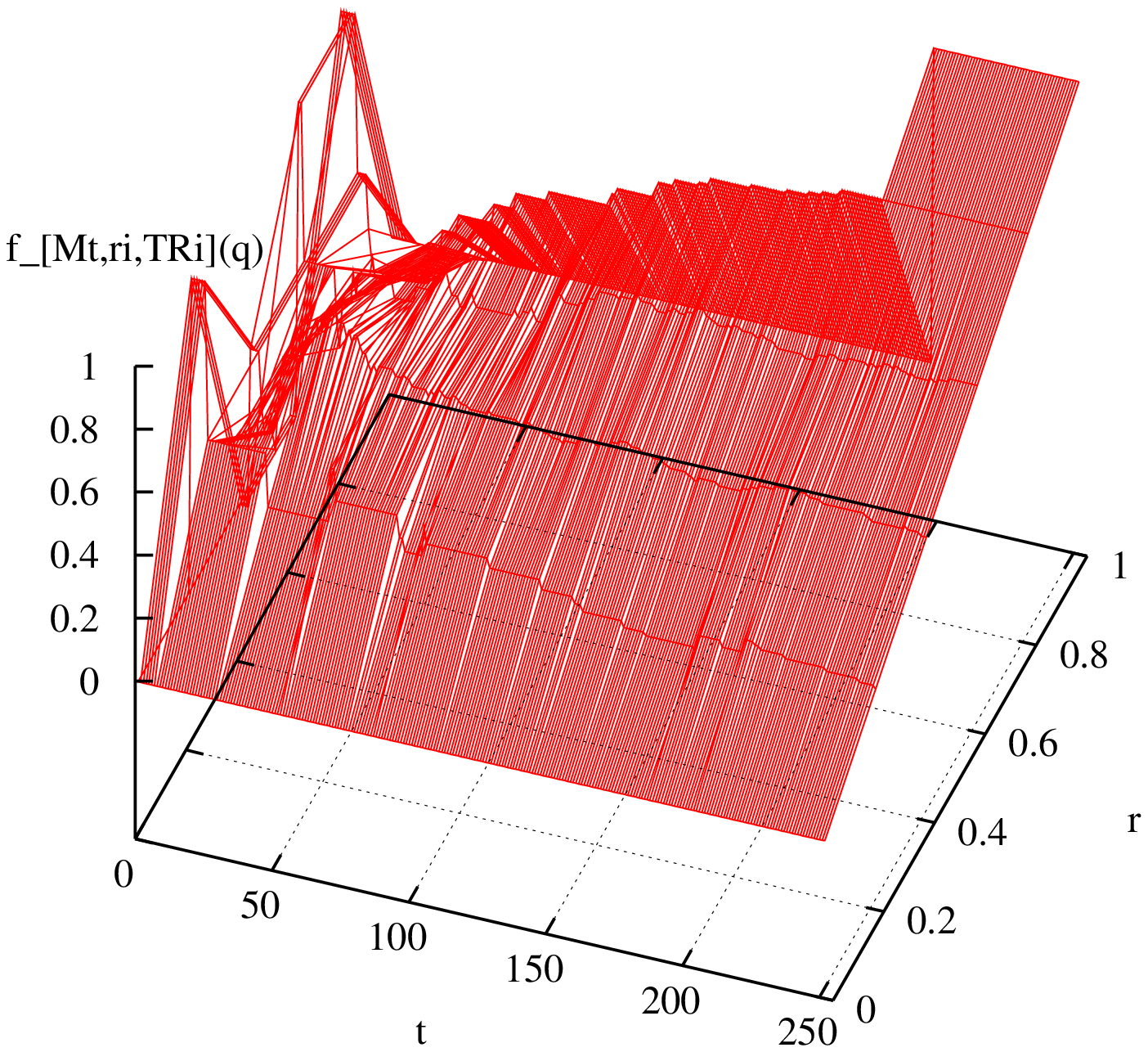,height=1.5in,clip=}\\
$q_1 \in TR_1$ & $q_6 \in TR_2$ \\
\psfig{file=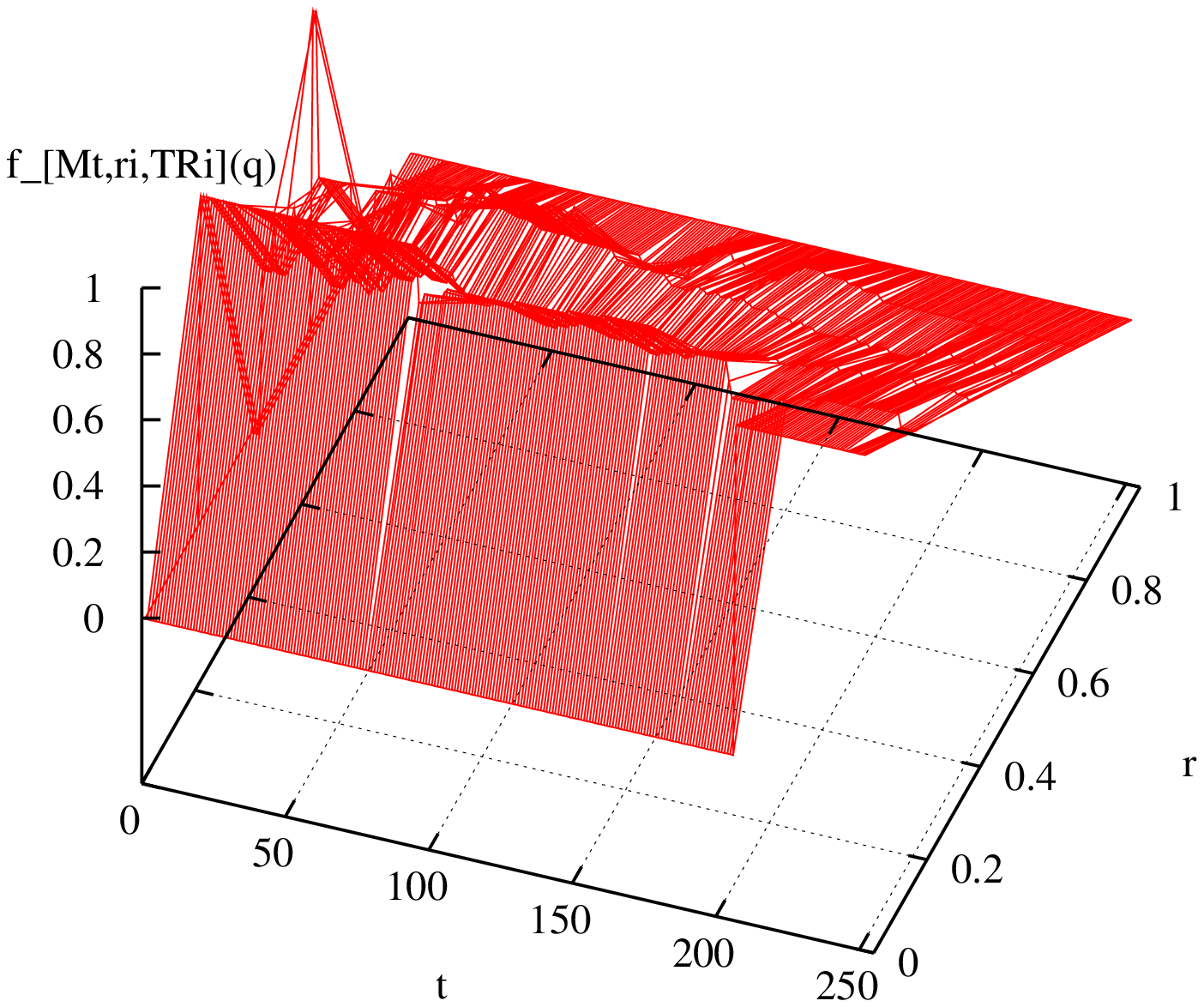,height=1.5in,clip=}&
\psfig{file=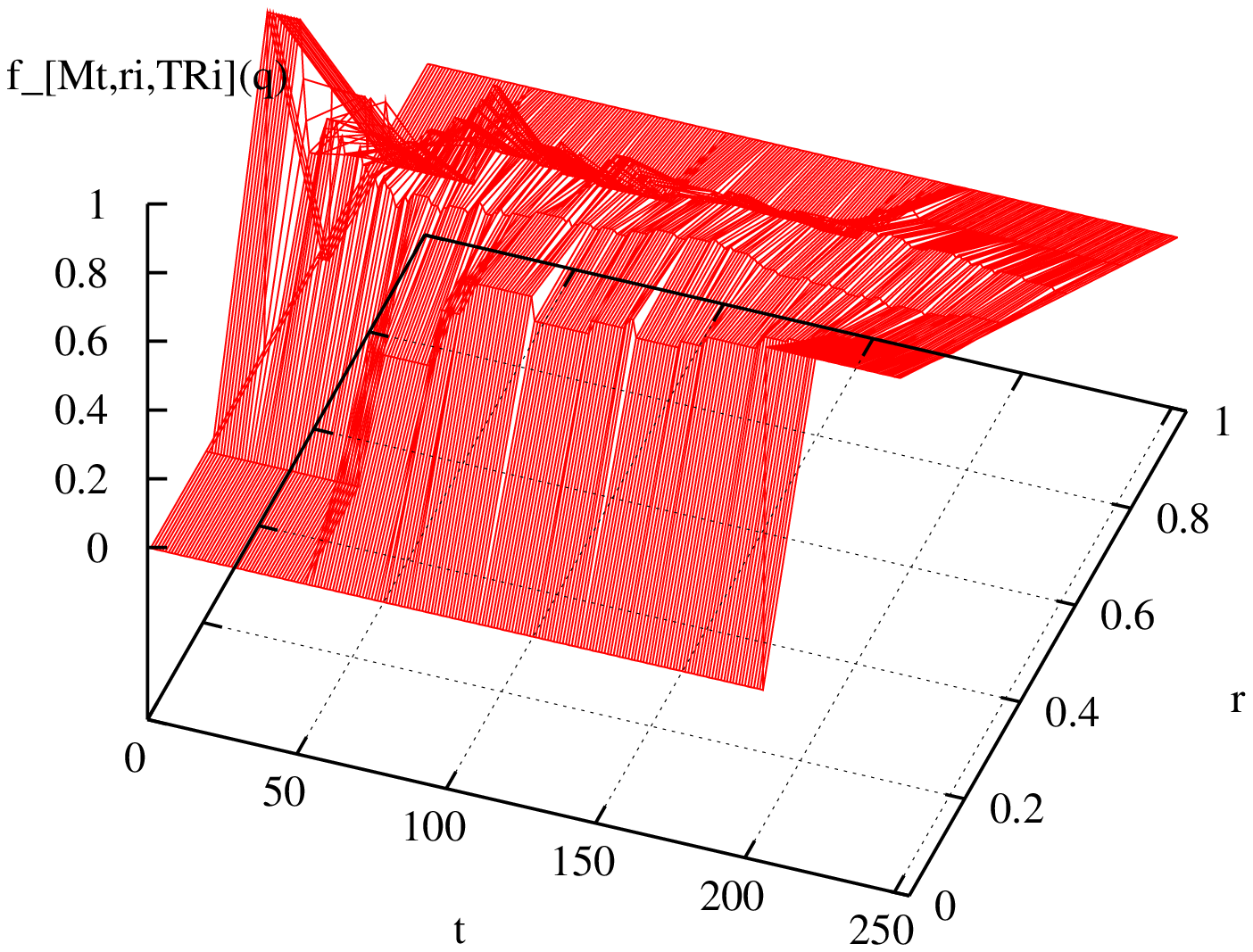,height=1.5in,clip=}\\
$q_2 \in TR_1$ & $q_7 \in TR_1$ \\
\psfig{file=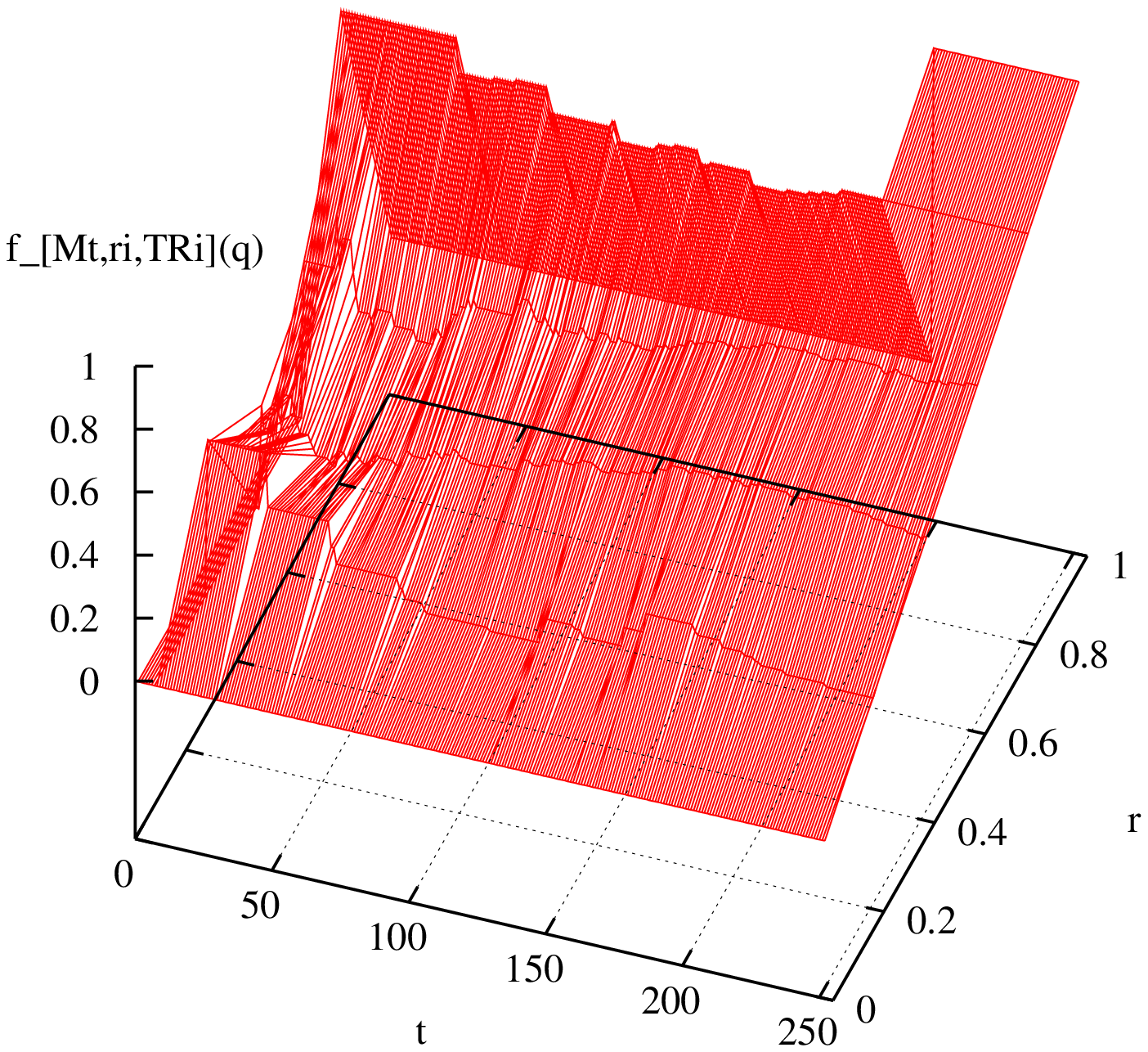,height=1.5in,clip=}&
\psfig{file=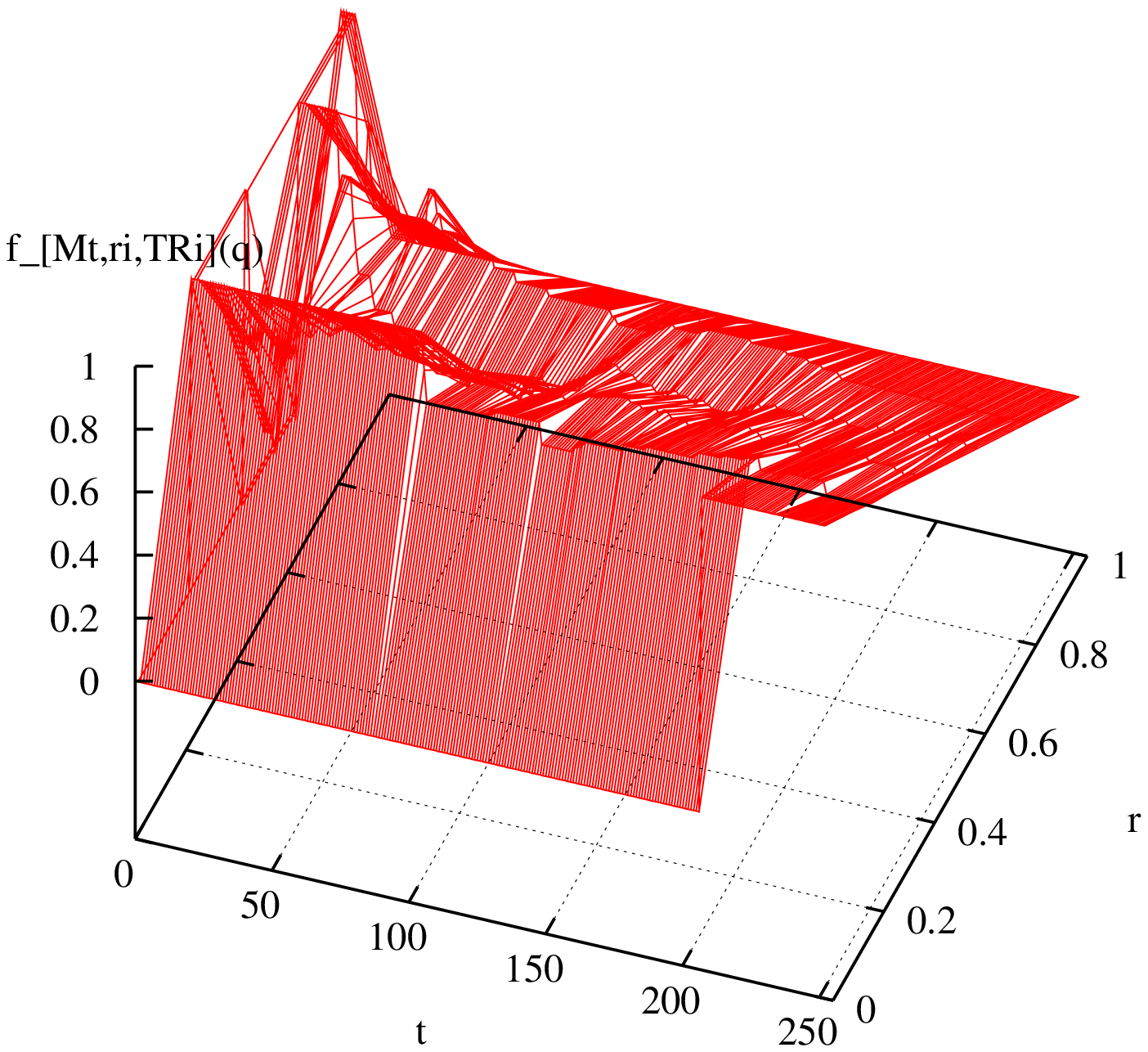,height=1.5in,clip=}\\
$q_3 \in TR_2$ & $q_8 \in TR_1$ \\
\psfig{file=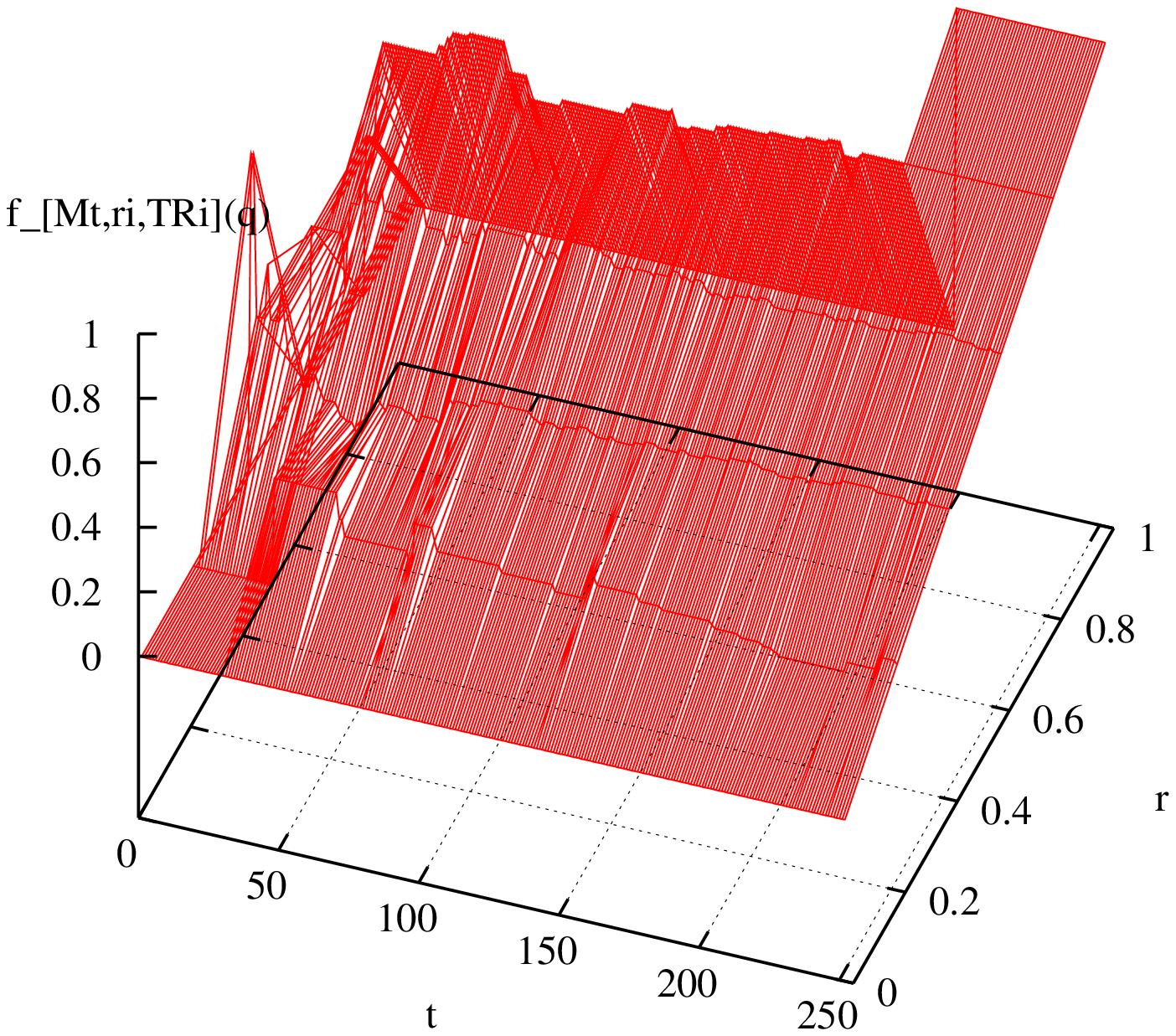,height=1.5in,clip=}&
\psfig{file=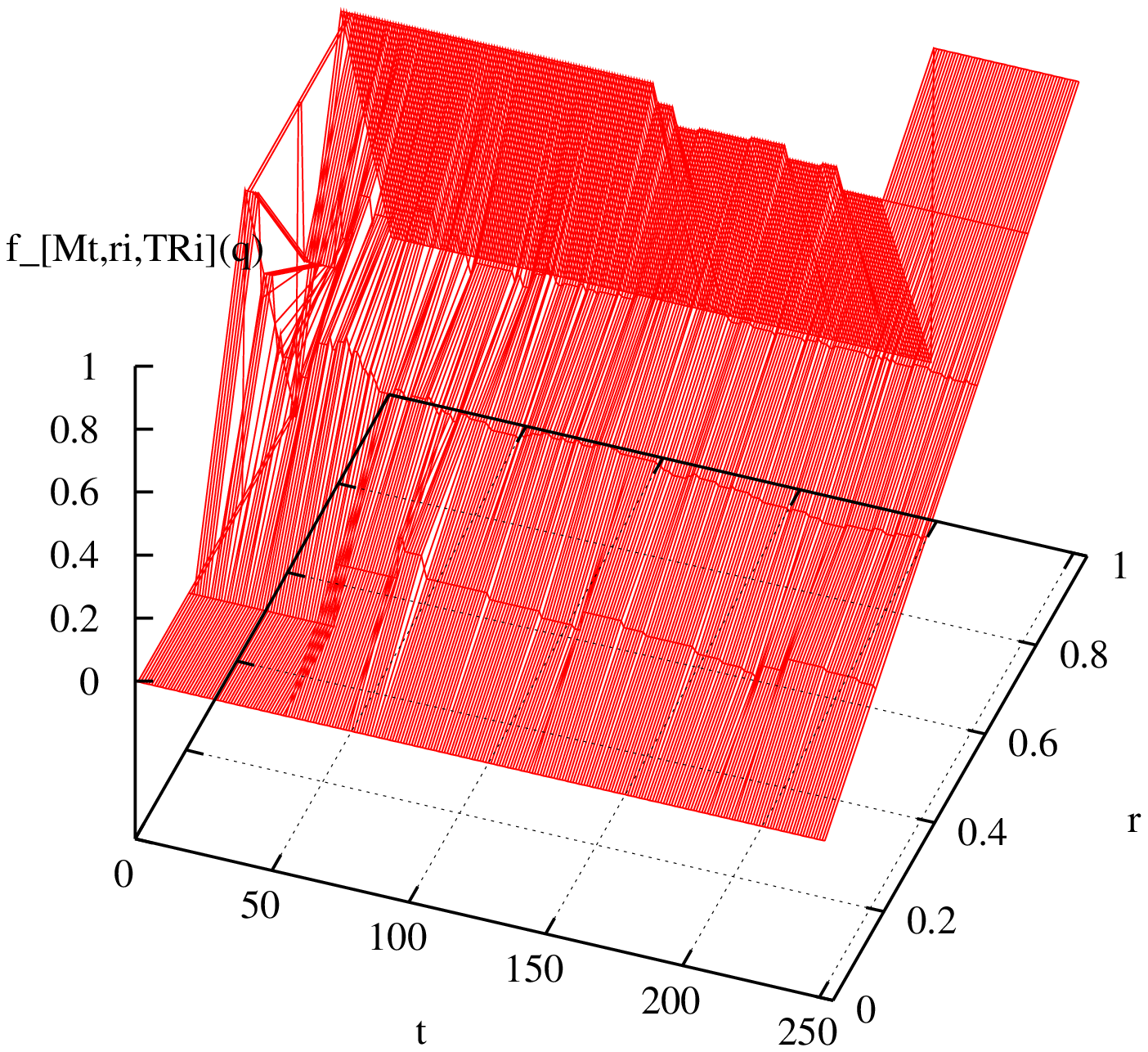,height=1.5in,clip=}\\
$q_4 \in TR_2$ & $q_9 \in TR_2$ \\
\end{tabular}
\caption{$f_{M_t,r_2,TR_2}(q)$ for each point $q$ and set $M_t$.
In each plot, the $x$ axis is $t$ that ranges from 0 to 252, the $y$
axis is $r$ that ranges from $0$ to $1$, and the $z$ axis is 
$f_{M_t,r_2,TR_2}$.}
\end{figure}

\clearpage
\newpage
\noindent{\large {\bf Appendix.}}

Listing of an awk script to calculate a model's ratings.
The script takes a file where each line contains the point indices of
the elements of a model (as in Table 1).

\vspace{-0.1in}
{\scriptsize
\begin{verbatim}
# <awk.rating>
# run:  awk -f awk.rating list_of_models > models.rating
# input:
# 3 5 6 8 9
# 0 1 2 6 8
# ...
BEGIN{
sizeTR1 = 5;
sizeTR2 = 5;
# full space = (x x x o o o o x x o)
c[0] = 1;  c[1] = 1;  c[2] = 1;  c[3] = 2;  c[4] = 2;  
c[5] = 2;  c[6] = 2;  c[7] = 1;  c[8] = 1;  c[9] = 2;
}
{ cover1 = 0;   
  cover2 = 0;   
  sizeM = NF;
  for (i=1; i<=NF; ++i) if (c[$i] == 1) cover1 ++; else cover2 ++;
  r1 = cover1/sizeTR1;
  r2 = cover2/sizeTR2;
  enr = r1 - r2;
  if (enr == 0.0) { xin = 0.0; xout = 0.0;}
  else {
     xin  = (1.0 - r2)/enr;
     xout = (0.0 - r2)/enr; 
  }
  printf ``model %s r1 %.2f r2 %.2f enr = r1-r2 = %.2f xin %.2f xout %.2f\n'',\
  $0,r1,r2,enr,xin,xout;
}
\end{verbatim}
}

Listing of an awk script to calculate the discriminant $Y_{12}$ for
each point $q$, given a list of models and their ratings (output of
awk.rating).

\vspace{-0.1in}
{\scriptsize
\begin{verbatim}
# <awk.discrim>
# run: awk -f awk.discrim models.rating 
# input:
# model 3 5 6 8 9  r1 0.20 r2 0.80 enr = r1-r2 = -0.60 xin -0.33 xout 1.33
# model 0 1 2 6 8  r1 0.80 r2 0.20 enr = r1-r2 = 0.60 xin 1.33 xout -0.33
# ...
{ 
  nm++;
  xin = $17;
  xout = $19;
#
# check for each point q, whether model m (this row) covers q,
# and increment y[q] accordingly
#
  for (q=0; q<10;++q) {
      inmodel = 0;
      for (i=2; i<=6; ++i) {
          if ($i == q) inmodel = 1;
      }
      if (inmodel) {
         y[q] = (y[q]*(nm-1) + xin)/nm; }
      else {      
         y[q] = (y[q]*(nm-1) + xout)/nm;}
  }
  printf ``Y(M,q)_(|M|= %3d ) ``, nm;
  for (q=0; q<10;++q) { printf ``%5.2f ``, y[q];} 
  printf ``\n'';
}
\end{verbatim}
}

\begin{table}[htbp]
{\small
\center
\caption{Models $m_t$ in $M_{0.5,A}$ in the order of $M = m_1, m_2,
..., m_{252}$.  Each model
is shown with its elements denoted by the indices $i$ of $q_i$ in $A$.
For example, $m_1 = \{q_3,q_5,q_6,q_8,q_9\}$. }
\begin{tabular}{||l|c||l|c||l|c||l|c||l|c||l|c||} \hline
$m_t$ & elements &
$m_t$ & elements &
$m_t$ & elements &
$m_t$ & elements &
$m_t$ & elements &
$m_t$ & elements \\ \hline
$m_{1}$&35689& $m_{43}$&12689& $m_{85}$&24578& $m_{127}$&01469& $m_{169}$&02468& $m_{211}$&02458 \\
$m_{2}$&01268& $m_{44}$&04569& $m_{86}$&23568& $m_{128}$&03679& $m_{170}$&35678& $m_{212}$&13457 \\
$m_{3}$&04789& $m_{45}$&01245& $m_{87}$&01267& $m_{129}$&04579& $m_{171}$&03589& $m_{213}$&24689 \\
$m_{4}$&25689& $m_{46}$&01458& $m_{88}$&01257& $m_{130}$&01237& $m_{172}$&34679& $m_{214}$&03478 \\
$m_{5}$&02679& $m_{47}$&15679& $m_{89}$&05679& $m_{131}$&24789& $m_{173}$&12346& $m_{215}$&23589 \\
$m_{6}$&34578& $m_{48}$&12457& $m_{90}$&24589& $m_{132}$&45689& $m_{174}$&12458& $m_{216}$&24679 \\
$m_{7}$&13459& $m_{49}$&02379& $m_{91}$&04589& $m_{133}$&16789& $m_{175}$&35789& $m_{217}$&02456 \\
\hline
\end{tabular}
}
\end{table}

\begin{table}[htbp]
\caption{Ratio of coverage of each point $q$ by members of $M_t$ as
$M_t$ expands.}
\begin{center}
{\tiny
\begin{tabular}{||l|cccccccccc|cccccccccc||}\hline
      & \multicolumn{10}{c|}{$N(M_t,q)$}
      & \multicolumn{10}{c||}{$Y(M_t,q)$} \\ \hline
$M_t$ & $q_0$ & $q_1$ & $q_2$ & $q_3$ & $q_4$
      & $q_5$ & $q_6$ & $q_7$ & $q_8$ & $q_9$
      & $q_0$ & $q_1$ & $q_2$ & $q_3$ & $q_4$
      & $q_5$ & $q_6$ & $q_7$ & $q_8$ & $q_9$ \\ \hline
$M_{1}$&0&0&0&1&0&1&1&0&1&1&0.00&0.00&0.00&1.00&0.00&1.00&1.00&0.00&1.00&1.00\\
$M_{2}$&1&1&1&1&0&1&2&0&2&1&0.50&0.50&0.50&0.50&0.00&0.50&1.00&0.00&1.00&0.50\\
$M_{3}$&2&1&1&1&1&1&2&1&3&2&0.67&0.33&0.33&0.33&0.33&0.33&0.67&0.33&1.00&0.67\\
$M_{4}$&2&1&2&1&1&2&3&1&4&3&0.50&0.25&0.50&0.25&0.25&0.50&0.75&0.25&1.00&0.75\\
$M_{5}$&3&1&3&1&1&2&4&2&4&4&0.60&0.20&0.60&0.20&0.20&0.40&0.80&0.40&0.80&0.80\\
$M_{6}$&3&1&3&2&2&3&4&3&5&4&0.50&0.17&0.50&0.33&0.33&0.50&0.67&0.50&0.83&0.67\\
\hline
\end{tabular}
}
\end{center}
\end{table}

\clearpage
\begin{table}[htbp]
\center\parbox{5in}{(Cont'd) Ratio of coverage of each point $q$ by
members of $M_t$ as $M_t$ expands.}
{\tiny
\begin{tabular}{||l|cccccccccc|cccccccccc||}\hline
      & \multicolumn{10}{c|}{$N(M_t,q)$}
      & \multicolumn{10}{c||}{$Y(M_t,q)$} \\ \hline
$M_t$ & $q_0$ & $q_1$ & $q_2$ & $q_3$ & $q_4$
      & $q_5$ & $q_6$ & $q_7$ & $q_8$ & $q_9$
      & $q_0$ & $q_1$ & $q_2$ & $q_3$ & $q_4$
      & $q_5$ & $q_6$ & $q_7$ & $q_8$ & $q_9$ \\ \hline
$M_{85}$&42&46&45&36&42&45&40&41&41&47&0.49&0.54&0.53&0.42&0.49&0.53&0.47&0.48&0.48&0.55\\
$M_{86}$&42&46&46&37&42&46&41&41&42&47&0.49&0.53&0.53&0.43&0.49&0.53&0.48&0.48&0.49&0.55\\
$M_{87}$&43&47&47&37&42&46&42&42&42&47&0.49&0.54&0.54&0.43&0.48&0.53&0.48&0.48&0.48&0.54\\
$M_{88}$&44&48&48&37&42&47&42&43&42&47&0.50&0.55&0.55&0.42&0.48&0.53&0.48&0.49&0.48&0.53\\
$M_{89}$&45&48&48&37&42&48&43&44&42&48&0.51&0.54&0.54&0.42&0.47&0.54&0.48&0.49&0.47&0.54\\
$M_{90}$&45&48&49&37&43&49&43&44&43&49&0.50&0.53&0.54&0.41&0.48&0.54&0.48&0.49&0.48&0.54\\
\hline
\end{tabular}
}
\end{table}

\clearpage
\begin{table}[htbp]
\center\parbox{5in}{(Cont'd) Ratio of coverage of each point $q$ by
members of $M_t$ as $M_t$ expands.}
{\tiny
\begin{tabular}{||l|cccccccccc|cccccccccc||}\hline
      & \multicolumn{10}{c|}{$N(M_t,q)$}
      & \multicolumn{10}{c||}{$Y(M_t,q)$} \\ \hline
$M_t$ & $q_0$ & $q_1$ & $q_2$ & $q_3$ & $q_4$
      & $q_5$ & $q_6$ & $q_7$ & $q_8$ & $q_9$
      & $q_0$ & $q_1$ & $q_2$ & $q_3$ & $q_4$
      & $q_5$ & $q_6$ & $q_7$ & $q_8$ & $q_9$ \\ \hline
$M_{169}$&86&84&84&85&85&81&86&83&81&90&0.51&0.50&0.50&0.50&0.50&0.48&0.51&0.49&0.48&0.53\\
$M_{170}$&86&84&84&86&85&82&87&84&82&90&0.51&0.49&0.49&0.51&0.50&0.48&0.51&0.49&0.48&0.53\\
$M_{171}$&87&84&84&87&85&83&87&84&83&91&0.51&0.49&0.49&0.51&0.50&0.49&0.51&0.49&0.49&0.53\\
$M_{172}$&87&84&84&88&86&83&88&85&83&92&0.51&0.49&0.49&0.51&0.50&0.48&0.51&0.49&0.48&0.53\\
$M_{173}$&87&85&85&89&87&83&89&85&83&92&0.50&0.49&0.49&0.51&0.50&0.48&0.51&0.49&0.48&0.53\\
$M_{174}$&87&86&86&89&88&84&89&85&84&92&0.50&0.49&0.49&0.51&0.51&0.48&0.51&0.49&0.48&0.53\\
\hline
\end{tabular}
}
\end{table}

\begin{table}[htbp]
\caption{Changes of $Y_{12}(q,M_t)$ as $M_t$ expands.  For each t, we show the
ratings for each new member of $M_t$, the values $X_{12}$ for this new
member, and $Y_{12}$ for the collection $M_t$ up to the inclusion of
this new member.}
\vspace{-0.15in}
\center
{\tiny
\begin{tabular}{||c|c|c|c|c|c|c|cccccccccc||}\hline
      &       &       &       &           & 
        \multicolumn{2}{c|}{$X_{12}(q,m_t)$ if} &
        \multicolumn{10}{c||}{$Y_{12}(q,M_t)$} \\ \cline{6-7}\cline{8-17}
$M_t$ & $m_t$ & $r_1$ & $r_2$ & $r_1-r_2$ & $q\in m_t$ & $q \not\in m_t$ 
      & $q_0$ & $q_1$ & $q_2$ & $q_3$ & $q_4$
      & $q_5$ & $q_6$ & $q_7$ & $q_8$ & $q_9$ \\ \hline
$M_{1}$&$m_{1}$&0.20&0.80&-0.60&-0.33&1.33  &1.33&1.33&1.33&-0.33&1.33&-0.33&-0.33&1.33&-0.33&-0.33\\
$M_{2}$&$m_{2}$&0.80&0.20&0.60&1.33&-0.33  &1.33&1.33&1.33&-0.33&0.50&-0.33&0.50&0.50&0.50&-0.33\\
$M_{3}$&$m_{3}$&0.60&0.40&0.20&3.00&-2.00  &1.89&0.22&0.22&-0.89&1.33&-0.89&-0.33&1.33&1.33&0.78\\
$M_{4}$&$m_{4}$&0.40&0.60&-0.20&-2.00&3.00  &2.17&0.92&-0.33&0.08&1.75&-1.17&-0.75&1.75&0.50&0.08\\
$M_{5}$&$m_{5}$&0.60&0.40&0.20&3.00&-2.00  &2.33&0.33&0.33&-0.33&1.00&-1.33&0.00&2.00&0.00&0.67\\
$M_{6}$&$m_{6}$&0.40&0.60&-0.20&-2.00&3.00  &2.44&0.78&0.78&-0.61&0.50&-1.44&0.50&1.33&-0.33&1.06\\
\hline
\end{tabular}
}
\end{table}

\clearpage
\begin{table}[htbp]
\center\parbox{5in}{(Cont'd) Changes of $Y_{12}(q,M_t)$ as $M_t$
expands.  For each t, we show the ratings for each new member of
$M_t$, the values $X_{12}$ for this new member, and $Y_{12}$ for the
collection $M_t$ up to the inclusion of this new member.}
\center
{\tiny
\begin{tabular}{||c|c|c|c|c|c|c|cccccccccc||}\hline
      &       &       &       &           & 
        \multicolumn{2}{c|}{$X_{12}(q,m_t)$ if} &
        \multicolumn{10}{c|}{$Y_{12}(q,M_t)$} \\ \cline{6-7}\cline{8-17}
$M_t$ & $m_t$ & $r_1$ & $r_2$ & $r_1-r_2$ & $q\in m_t$ & $q \not\in m_t$ 
      & $q_0$ & $q_1$ & $q_2$ & $q_3$ & $q_4$
      & $q_5$ & $q_6$ & $q_7$ & $q_8$ & $q_9$ \\ \hline
$M_{85}$&$m_{85}$&0.60&0.40&0.20&3.00&-2.00  &1.18&1.02&0.69&-0.16&0.00&-0.29&0.39&0.84&1.27&0.06\\
$M_{86}$&$m_{86}$&0.40&0.60&-0.20&-2.00&3.00  &1.20&1.04&0.65&-0.18&0.04&-0.31&0.36&0.87&1.24&0.09\\
$M_{87}$&$m_{87}$&0.80&0.20&0.60&1.33&-0.33  &1.20&1.05&0.66&-0.18&0.03&-0.31&0.38&0.87&1.22&0.09\\
$M_{88}$&$m_{88}$&0.80&0.20&0.60&1.33&-0.33  &1.20&1.05&0.67&-0.18&0.03&-0.29&0.37&0.88&1.20&0.08\\
$M_{89}$&$m_{89}$&0.40&0.60&-0.20&-2.00&3.00  &1.16&1.07&0.70&-0.15&0.06&-0.31&0.34&0.85&1.22&0.06\\
$M_{90}$&$m_{90}$&0.40&0.60&-0.20&-2.00&3.00  &1.18&1.09&0.67&-0.11&0.04&-0.33&0.37&0.87&1.18&0.04\\
\hline
\end{tabular}
}
\end{table}

\clearpage
\begin{table}[htbp]
\center\parbox{5in}{(Cont'd) Changes of $Y_{12}(q,M_t)$ as $M_t$
expands.  For each t, we show the ratings for each new member of
$M_t$, the values $X_{12}$ for this new member, and $Y_{12}$ for the
collection $M_t$ up to the inclusion of this new member.}
\center
{\tiny
\begin{tabular}{||c|c|c|c|c|c|c|cccccccccc||}\hline
      &       &       &       &           & 
        \multicolumn{2}{c|}{$X_{12}(q,m_t)$ if} &
        \multicolumn{10}{c|}{$Y_{12}(q,M_t)$} \\ \cline{6-7}\cline{8-17}
$M_t$ & $m_t$ & $r_1$ & $r_2$ & $r_1-r_2$ & $q\in m_t$ & $q \not\in m_t$ 
      & $q_0$ & $q_1$ & $q_2$ & $q_3$ & $q_4$
      & $q_5$ & $q_6$ & $q_7$ & $q_8$ & $q_9$ \\ \hline
$M_{169}$&$m_{169}$&0.60&0.40&0.20&3.00&-2.00  &1.07&1.05&0.95&0.19&-0.01&-0.22&0.06&0.88&1.06&-0.02\\
$M_{170}$&$m_{170}$&0.40&0.60&-0.20&-2.00&3.00  &1.08&1.06&0.96&0.18&0.01&-0.23&0.05&0.86&1.04&0.00\\
$M_{171}$&$m_{171}$&0.40&0.60&-0.20&-2.00&3.00  &1.06&1.07&0.97&0.16&0.03&-0.25&0.07&0.87&1.02&-0.01\\
$M_{172}$&$m_{172}$&0.20&0.80&-0.60&-0.33&1.33  &1.06&1.07&0.97&0.16&0.03&-0.24&0.06&0.87&1.02&-0.01\\
$M_{173}$&$m_{173}$&0.40&0.60&-0.20&-2.00&3.00  &1.07&1.05&0.96&0.15&0.01&-0.22&0.05&0.88&1.03&0.00\\
$M_{174}$&$m_{174}$&0.60&0.40&0.20&3.00&-2.00  &1.06&1.06&0.97&0.14&0.03&-0.20&0.04&0.86&1.05&-0.01\\
\hline
\end{tabular}
}
\end{table}

\end{document}